\documentclass[lettersize,journal]{IEEEtran}
\usepackage{amsmath,amsfonts}
\usepackage{algorithmic}
\usepackage{algorithm}
\usepackage{array}
\usepackage[caption=false,font=normalsize,labelfont=sf,textfont=sf]{subfig}
\usepackage{textcomp}
\usepackage{stfloats}
\usepackage{url}
\usepackage{verbatim}
\usepackage{graphicx}
\usepackage{cite}
\usepackage{abbreviation}
\usepackage[colorlinks=true, linkcolor=blue, citecolor=blue, urlcolor=blue]{hyperref}
\usepackage[table]{xcolor}  
\usepackage{amsmath}
\usepackage{amssymb}
\usepackage{makecell}       
\usepackage{adjustbox}
\usepackage{graphicx}       
\usepackage{multirow}       
\usepackage{booktabs}       
\usepackage[table]{xcolor}
\usepackage{colortbl}
\usepackage{threeparttable}
\usepackage{pifont}

\definecolor{deepred}{RGB}{139,0,0}
\definecolor{deepgreen}{RGB}{0,100,0}

\newcommand{\cmark}{\ding{51}}
\newcommand{\xmark}{\ding{55}}

\begin{document}

\title{Hypergraph as Language}

\author{Mengqi~Lei, 
        Guohuan~Xie,
        Shihui~Ying,~\IEEEmembership{Senior Member,~IEEE},
        Shaoyi~Du,~\IEEEmembership{Senior Member,~IEEE},
        Jun-Hai~Yong,
        Chuan Shi,~\IEEEmembership{Senior Member,~IEEE},
        Ling Tian,
        Siqi~Li,
        Yue~Gao,~\IEEEmembership{Senior Member,~IEEE}\vspace{-0.2cm}
\IEEEcompsocitemizethanks{
\IEEEcompsocthanksitem Mengqi Lei, Guohuan Xie, Jun-Hai Yong, Siqi Li, and Yue Gao are with \{BNRist, THUIBCS, BLBCI, School of Software\}, Tsinghua University, Beijing, 100084, China. (leimq25@mails.tsinghua.edu.cn, stuxiemol@gmail.com, yongjh@tsinghua.edu.cn, lisiqi19971013@gmail.com, kevin.gaoy@gmail.com)
\IEEEcompsocthanksitem Shihui Ying is with the Shanghai Institute of Applied Mathematics and Mechanics, School of Mechanics and Engineering Science, Shanghai University, Shanghai 200072, China. (shying@shu.edu.cn)
\IEEEcompsocthanksitem Shaoyi Du is with the State Key Laboratory of Human-Machine Hybrid Augmented Intelligence, National Engineering Research Center for Visual Information and Applications, and Institute of Artificial Intelligence and Robotics, Xi'an Jiaotong University, Xi'an, 710049, China. (dushaoyi@xjtu.edu.cn)
\IEEEcompsocthanksitem Chuan Shi is with the Beijing Key Lab of Intelligent Telecommunications Software and Multimedia, Beijing University of Posts and Telecommunications, Beijing 100876, China. (shichuan@bupt.edu.cn)
\IEEEcompsocthanksitem Ling Tian is with the School of Computer Science and Engineering, University of Electronic Science and Technology of China, Chengdu, 611731, China. (lingtian@uestc.edu.cn)
}
}

\markboth{Hypergraph as Language}%
{Lei \MakeLowercase{\textit{et al.}}: Hypergraph as Language}


\maketitle

\begin{abstract}
Large language models (LLMs) have recently demonstrated substantial potential in modeling relational structures. However, existing approaches remain fundamentally graph-centric: they primarily focus on processing pairwise graph structures into tokens that LLMs can understand. In contrast, many real-world relational patterns do not naturally conform to the pairwise-edge assumption, and are better modeled as high-order associations in hypergraphs. When facing hypergraph structures, existing methods often fail to preserve the native semantics that multiple objects are jointly connected by the same high-order relation, thereby limiting their ability to effectively exploit complex structures.
To address this limitation, we put forward the \emph{Hypergraph as Language} perspective and propose Hyper-Align, a hypergraph-native alignment framework for large language models.
Hyper-Align compiles the query-object-centered hypergraph context into a sequence of hypergraph tokens that can be directly consumed by a base LLM. Specifically, we first introduce Hypergraph Incidence Detail Template with Overview (HIDT-O), which serializes high-order association structures into a fixed-shape hybrid template combining local incidence details and overview-level summaries. We then design a Hypergraph Incidence Projector (HIP), which maps native high-order incidence structures into the LLM token space through explicit semantic-structural decoupling and bidirectional message passing between vertices and hyperedges. Building on this, we further define a concrete Hypergraph-as-Language input protocol, which jointly feeds hypergraph tokens and textual prompts into a frozen base LLM, thereby supporting both vertex-level and hyperedge-level tasks under a unified question-answering paradigm. Furthermore, to systematically evaluate the capability of different methods in hypergraph structural modeling, we introduce the HyperAlign-Bench. Extensive experiments show that our Hyper-Align significantly outperforms existing methods across in-domain and zero-shot evaluations. The code is available at: \texttt{\url{https://github.com/Mengqi-Lei/Hypergraph-as-Language}}.
\end{abstract}

\begin{figure}[t]
    \centering
    \includegraphics[width=0.9\linewidth]{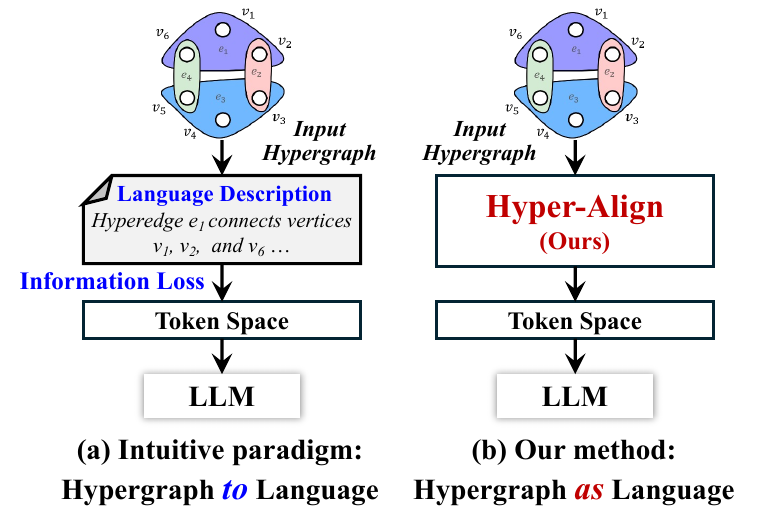}
    \vspace{-12pt}
    \caption{Illustration of an intuitive text-based Hypergraph-\textbf{to}-Language paradigm and our Hypergraph-\textbf{as}-Language paradigm. In the former paradigm, high-order structural details undergo severe compression and loss, whereas the latter paradigm (\ie, our approach) effectively preserves and encodes the higher-order structures that are widespread in the real-world data.}
    \label{fig:intro}
\end{figure}

\section{Introduction}

\IEEEPARstart{L}{arge} language models (LLMs) have demonstrated strong capabilities in language understanding, knowledge transfer, and unified task modeling, which has further accelerated their extension to structured data domains \cite{llm_1,llm_2,llm_3,llm_4}. In recent years, LLM-based research on graph-structured data has gradually become an active research topic, mainly following two paradigms \cite{llm_graph_survey,llm_graph_survey2,llm_graph_survey3}. One line of methods rewrites graph structures into natural-language or code-style descriptions \cite{l_is_g_need,graphllm, codegraph, graph_as_code_2}, enabling LLMs to directly perform graph tasks in the textual space. The other line of methods transforms graph structures into sequences of graph tokens that are compatible with the input space of LLMs, using a frozen or nearly frozen LLM as a unified interface for various tasks and even zero-shot inference \cite{graphgpt,llaga,tea_glm,promptgfm}. Although these methods have significantly advanced the integration of LLMs and graph learning, their modeling assumptions remain fundamentally graph-centric: the basic structural units are typically pairwise adjacency, node neighborhoods, or token sequences derived from graphs.

However, many high-order relations in the real world do not naturally conform to the pairwise-edge assumption of ordinary graphs \cite{hg_old_book,hg_learning,hg_book,beyond_pair}. Data such as paper co-citation, group interactions, and multi-entity collaboration are better modeled as hypergraphs, where a hyperedge can simultaneously connect an arbitrary number of vertices. Accordingly, the semantic focus is no longer whether two vertices are connected, but how a set of objects are associated as a whole. This means that the native structural unit of a hypergraph is not pairwise adjacency, but the high-order association established between vertices and hyperedges \cite{hgnn,hypergcn}. In this regard, directly applying existing graph-centric methods requires expanding a hypergraph into multiple pairwise edges (\eg, via clique expansion), which inevitably loses the high-order association information in the original hypergraph \cite{allset,hg_isomorph}.

Recently, several preliminary efforts have emerged to combine hypergraphs with large language models. LLMHG \cite{llmhg}, HeLLM \cite{hellm}, and Hyper-RAG \cite{hyperrag} respectively focus on recommendation systems, multimodal recommendation, and retrieval-augmented generation (RAG) scenarios, while HyperLLM \cite{hyperllm} focuses on leveraging LLMs to generate hypergraphs from textual data.
LLM4Hypergraph \cite{llm4hg} built a systematic benchmark for hypergraph understanding, aiming to explore converting hypergraphs into natural language so that large models can understand them. We call this rather intuitive paradigm ``Hypergraph to Language'', whose serialized text-based descriptions can compress or omit high-order structural details severely.
Despite these efforts, a fundamental question remains largely unexplored: \textbf{Can we make hypergraphs directly understandable by LLMs as a language-like structural input, so that an LLM can natively model high-order associations and uniformly handle hypergraph tasks?} We refer to this new line of research as ``Hypergraph as Language'', as shown in Fig.~\ref{fig:intro}.

Based on the \emph{hypergraph as language} view, we propose Hyper-Align, the first hypergraph-native alignment framework for large language models.
Hyper-Align compiles the query-object-centered high-order association structure into a sequence of continuous hypergraph tokens that can be directly consumed by a base LLM, and performs inference under a unified question-answering paradigm. 
Specifically, we first propose a hypergraph-native serialization approach, Hypergraph Incidence Detail Template with Overview (HIDT-O). From the vertex-hyperedge incidence perspective, it serializes high-order association structures into a fixed-shape template comprising local incidence details and overview-level summaries. 
Next, at the representation alignment level, we design the Hypergraph Incidence Projector (HIP). Unlike the shared-MLP-style projectors commonly used in existing graph-LLM methods, our HIP explicitly decouples semantics and structure, distinguishing different roles such as vertices, hyperedges, and overview components. Moreover, it performs a high-order bidirectional message passing between vertices and hyperedges within the projector, thereby mapping the native high-order association structures into the LLM token space. 
Building on this, we further define a concrete Hypergraph-as-Language input protocol, which uses a three-part Background-Details-Question prompt to jointly feed hypergraph tokens and textual context into a frozen base LLM. Finally, in hypergraph alignment tuning, we design two auxiliary supervision tasks, namely order bucket reconstruction and relation reconstruction, to optimize the parameters of the HIP jointly with the main task loss. Notably, Hyper-Align is not limited to hypergraphs but naturally extends to ordinary graphs, since an ordinary graph can be viewed as a special hypergraph where each hyperedge associates exactly two vertices.
To systematically evaluate the capability of different methods in high-order association modeling, we further construct a HyperAlign-Bench, which contains two core tasks, vertex classification and hyperedge classification, and supports in-domain and zero-shot evaluations.

In summary, the contributions of this paper can be summarized as follows.

\begin{itemize}
\vspace{-6pt}
    \item We introduce the \emph{Hypergraph as Language} perspective and propose Hyper-Align, the first hypergraph-native alignment framework for LLMs. 
    \item We propose HIDT-O, which serializes high-order association structures into a hybrid template of local details and overview-level summaries from the vertex-hyperedge incidence perspective. 
    \item We propose HIP and define a concrete Hypergraph-as-Language input protocol, establishing a unified alignment interface among high-order incidence structures, textual contexts, and LLMs. Meanwhile, we design auxiliary supervision in hypergraph alignment tuning to jointly optimize HIP parameters. 
    \item We construct HyperAlign-Bench, a fair and reproducible benchmark for high-order association modeling. Extensive experiments show that Hyper-Align significantly outperforms existing methods in both in-domain and zero-shot evaluations, verifying the necessity and effectiveness of the proposed method. 
\end{itemize}

\section{Related Work}

\subsection{LLMs for Graph Structural Data}

Existing studies on how to enable LLMs to understand and process graph-structured data can be broadly categorized into two paradigms \cite{llm_graph_survey,llm_graph_survey2,llm_graph_survey3}. The earlier paradigm follows the graph-as-text (or graph-as-code) route, whose core idea is to rewrite graph structures into natural-language descriptions or code-style representations, and then enable LLMs to perform graph tasks in the textual space through instruction tuning or in-context learning \cite{l_is_g_need,graphllm,codegraph,graph_as_code_2}. InstructGLM directly describes multi-scale graph structures in natural language, promoting early explorations in generative graph learning \cite{l_is_g_need}. GraphLLM further points out that the conventional Graph2Text process itself may become a bottleneck, and attempts to enhance the graph reasoning capability of LLMs through end-to-end structural modeling \cite{graphllm}.

A more recent mainstream paradigm is the graph-to-token route \cite{graphgpt,llaga,tea_glm,promptgfm,unigraph,gofa}. The core question in this direction is not how to write graphs as text, but how to transform graph structures into graph tokens that are compatible with the input space of LLMs. GraphGPT injects graph structural knowledge into LLMs through graph instruction tuning \cite{graphgpt}. LLaGA maps node sequences in graphs into the LLM token space through structure-aware templates and a lightweight projector \cite{llaga}. TEA-GLM emphasizes the explicit alignment between graph neural network representations and LLM token embeddings, aiming to improve zero-shot generalization across datasets and tasks \cite{tea_glm}. PromptGFM further proposes a language-based graph vocabulary, unifying graph structure understanding and reasoning \cite{promptgfm}.

Although the above graph-to-token route has demonstrated that aligning structures to language models is a highly promising technical direction, their shared premise remains the ordinary graph: the basic structural units being modeled are nodes, binary edges, and their local neighborhoods. For many real-world relational data, this premise is insufficient \cite{hg_old_book,hg_learning,hg_book,beyond_pair}. The semantic core of phenomena such as co-occurrence, group interactions, and multi-agent events lies in the holistic fact that a set of objects are jointly associated by the same high-order relation. Existing studies have shown that graphizing a hypergraph flattens the group-level identity information within hyperedges, and is therefore insufficient to preserve the native semantics of high-order relations \cite{allset,hg_isomorph}. Therefore, although existing graph-LLM methods provide an important foundation for structure-language alignment, they are still insufficient to directly address high-order association modeling in hypergraph scenarios.

\vspace{-12pt}

\subsection{Hypergraph Learning and Preliminary Explorations of Hypergraph-LLMs }

Unlike the pairwise connections modeled by ordinary graphs, a hyperedge in a hypergraph can simultaneously connect an arbitrary number of vertices, and is therefore better suited for capturing high-order associations that widely exist in real-world data \cite{hg_old_book,hg_learning,hg_book,beyond_pair}. Centered on this structural property, hypergraph learning has become a key technique for studying associative relationships \cite{hgnn,hypergcn,hg_survey,hg_survey2}. Early work such as Hypergraph Neural Network (HGNN) directly takes hyperedges as the basic structural units for message passing, initiating the study of hypergraph neural networks \cite{hgnn}. Subsequently, Hyper-SAGNN further leverages self-attention mechanisms to handle hyperedges of variable sizes and supports high-order relation prediction \cite{hypersagnn}. AllSet unifies hypergraph neural networks from the perspective of multiset functions, emphasizing the core inductive bias that ``a hyperedge is a set rather than a list of edges'' \cite{allset}.

In recent years, researchers have further begun to explore the joint modeling of high-order structures and textual semantics. HyperBERT incorporates hypergraph-aware layers into pretrained language models for vertex classification on text-attributed hypergraphs \cite{hyperbert}. Hyper-FM investigates the scalability of hypergraph neural networks from the perspective of foundation models \cite{hyperfm}. DHG-Bench systematically compares different categories of hypergraph neural network methods across multiple tasks and datasets from the benchmark perspective \cite{dhgbench}. Although these works have started to study the modeling of text-attributed hypergraphs, their core approach remains an HGNN encoder. In most cases, textual attributes are merely converted by text encoders such as BERT into features that can be processed by HGNNs.

Meanwhile, in the past two years, several preliminary efforts have also emerged to combine hypergraphs with large language models. LLM4Hypergraph constructs a systematic benchmark for hypergraph understanding and reasoning, and designs multiple prompting strategies to analyze whether LLMs are capable of handling hypergraphs \cite{llm4hg}. Beyond this, most existing studies still follow scenario-specific integration. Methods such as LLMHG and HeLLM mainly target recommendation systems \cite{llmhg,hellm}, Hyper-RAG focuses on retrieval augmentation over complex relational knowledge \cite{hyperrag}, and HyperLLM aims to extract high-order association structures from text using large language models \cite{hyperllm}. These studies indicate that hypergraphs can indeed bring high-order structural benefits to large language models. However, their focus is mainly on benchmarks, prompting, or specific applications, rather than on establishing a unified hypergraph-language alignment framework.
More importantly, existing attempts largely follow what we term the ``Hypergraph to Language'' paradigm, where hypergraph structure is first translated into natural-language descriptions and then fed to the LLM. 
%
Although intuitive, this paradigm provides an indirect structural interface. Under finite context budgets, textual descriptions can compress or truncate the original hypergraph, while typed vertex-hyperedge incidence is not exposed as explicitly as in a structural representation. Consequently, hyperedge identity and group membership can be severely weakened during serialization.
In contrast, our ``Hypergraph as Language'' view does not describe a hypergraph only through textual narration; instead, it directly organizes native hypergraph structure into structural tokens that can be consumed by the LLM. Since the incidence representation itself is lossless with respect to the original vertex-hyperedge relations, this formulation provides a principled path toward structure-preserving hypergraph-language alignment.

In summary, existing hypergraph learning methods mainly study how to achieve better association modeling on hypergraphs, while existing hypergraph-LLM studies mainly investigate whether LLMs can understand hypergraphs or how to combine hypergraphs with LLMs in specific scenarios. However, they have not yet answered a more fundamental question: 
Can we make hypergraphs directly understandable by LLMs as a language-like structural input, so that an LLM can natively model high-order associations and uniformly handle hypergraph tasks? The proposed Hyper-Align is designed to fill this gap: we make the first attempt to construct a hypergraph-native alignment framework for large language models.

\section{Hyper-Align Framework}
\label{sec:theory}

\subsection{Problem Formulation}
Given a hypergraph, we denote it as:
\begin{equation}
    \mathcal{H}=(\mathcal{V},\mathcal{E},\mathcal{X},\mathcal{Z})
\end{equation}
where $\mathcal{V}$ denotes the set of vertices, 
$\mathcal{E}=\{e_1,\ldots,e_M\}$ denotes the set of hyperedges, 
and each hyperedge $e_i\subseteq\mathcal{V}$ is a subset of vertices that may contain an arbitrary number of vertices.
Here, $\mathcal{X}=\{x_v\mid v\in\mathcal{V}\}$ represents the textual attributes of vertices, while $\mathcal{Z}=\{z_e\mid e\in\mathcal{E}\}$ represents optional textual attributes or metadata associated with hyperedges. 
We define the vertex degree and the hyperedge degree respectively as:
$d(v)=|\{e\in\mathcal{E}\mid v\in e\}|,\quad r(e)=|e|$,
where $r(e)$ is also used as the hyperedge order in our order bucket notation in Sec.~\ref{sec:o_suffix}.
We further define the incidence matrix as:
$B\in\{0,1\}^{|\mathcal{V}|\times |\mathcal{E}|},$
where each entry $B_{v,e}$ indicates whether vertex $v$ is contained in hyperedge $e$: 
$B_{v,e}=1$ if $v\in e$, and $B_{v,e}=0$ otherwise.
Hyper-Align adopts a unified object-centric interface. Given a query center
$c\in\mathcal{V}\cup\mathcal{E}$,
the objective of the model is to map the high-order relational context centered at $c$ into a continuous token sequence that can be directly consumed by an LLM, and then enable a frozen LLM to perform downstream prediction tasks under a unified question-answering paradigm. Notably, when all hyperedges satisfy $r(e)=2$, this formulation naturally degenerates to the ordinary graph setting.

\begin{figure*}[t]
    \centering
    \includegraphics[width=1\linewidth]{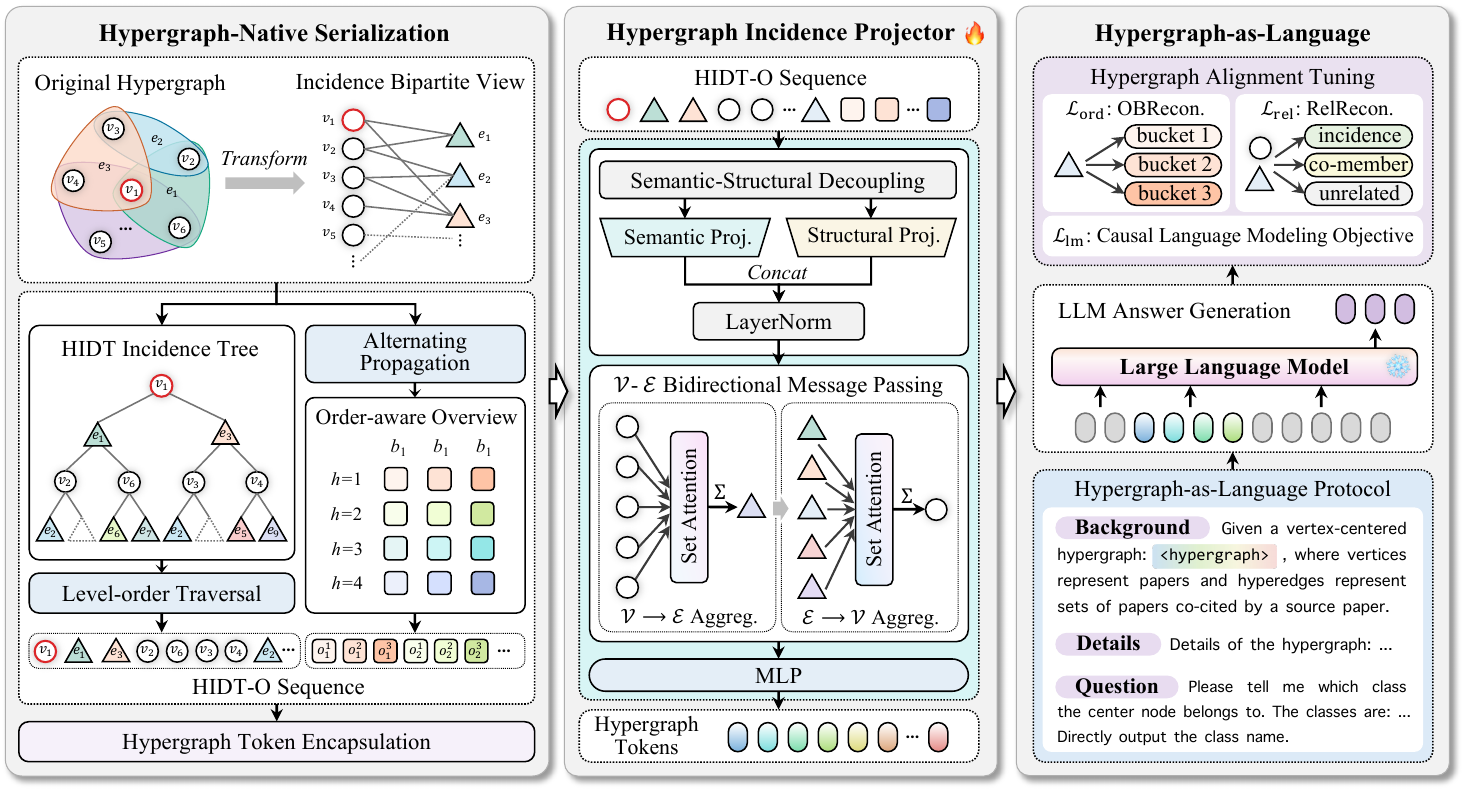}
    \vspace{-16pt}
    \caption{
        Overall framework of the proposed Hyper-Align. In the figure, ``OBRecon.'' and ``RelRecon.'' denote order bucket reconstruction and relation reconstruction, respectively. Circles and triangles denote vertices and hyperedges, respectively, and gray dashed slots denote padded slots.
}
    \label{fig:frame}
\end{figure*}

\subsection{Overall Framework}

As shown in Fig.~\ref{fig:frame}, for an arbitrary center object $c$, the overall computational pipeline of the proposed Hyper-Align can be written as:
\begin{equation}
    c \xrightarrow{\;\text{HIDT-O}\;} \Pi(c)
\xrightarrow{\;\mathrm{HIP}_{\phi}\;}
T(c)\in\mathbb{R}^{L_\mathcal{H}\times d_{\mathrm{llm}}}
\xrightarrow{\;\mathrm{LLM}_{\Theta}\;} y,
\end{equation}
where $\Pi(c)=(u_1,\ldots,u_L)$ is a fixed-length structural sequence generated by the hypergraph-native serialization template, $\mathrm{HIP}_{\phi}$ denotes the proposed Hypergraph Incidence Projector (HIP), $T(c)$ is a sequence of hypergraph tokens of length $L_\mathcal{H}$, with the same dimensionality as the hidden space of the backbone LLM, and $\mathrm{LLM}_{\Theta}$ denotes a frozen large language model.
The overall framework consists of three mutually coupled components:
(1) \textbf{HIDT-O}, which serializes the hypergraph structure into a fixed-length token sequence;
(2) \textbf{HIP}, which aligns the semantic and structural information in the sequence with the LLM token space;
(3) the \textbf{Hypergraph-as-Language protocol}, which feeds the hypergraph tokens together with natural-language context into the frozen LLM through a three-part Background-Details-Question prompt.
During the entire hypergraph alignment tuning process, only the parameters of HIP are updated. No task-specific head is introduced, and the backbone LLM remains unchanged.

\subsection{Hypergraph-Native Serialization: HIDT-O}

\subsubsection{Hypergraph Incidence Detail Template}
\begin{figure}[t]
    \centering
    \includegraphics[width=1\linewidth]{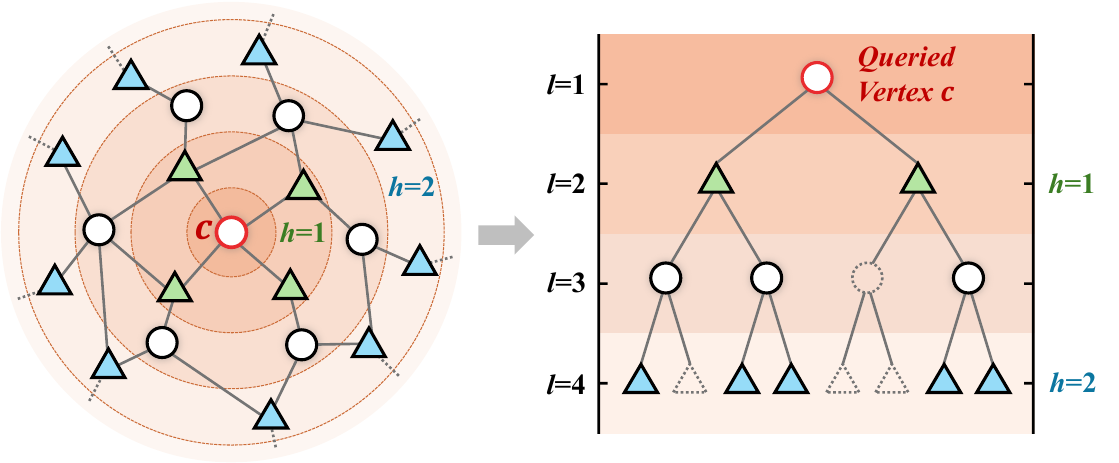}
    \caption{Illustration of HIDT. The left part shows a query-centered local hypergraph context, while the right part shows the corresponding fixed-shape HIDT incidence tree. Here, $h$ indicates the hyperedge-hop shell, $l$ denotes the template layer.}
    \label{fig:hidt}
\end{figure}

To compile high-order association structures into fixed-length sequences, we propose the Hypergraph Incidence Detail Template (HIDT). As shown in Fig.~\ref{fig:hidt}, given a center object $c$, HIDT constructs a fixed-shape incidence tree from the vertex-hyperedge incidence bipartite perspective, where vertex layers and hyperedge layers strictly alternate. 
It is worth noting that this incidence bipartite representation is lossless with respect to the vertex-hyperedge incidence structure, and can fully retain the structural information of the original hypergraph \cite{hgnn_math_book}.

Specifically, we place the center object $c$ at layer 0 as the root node. When $c\in\mathcal{V}$, the first layer samples a fixed number of hyperedges from the incident hyperedges of $c$. The second layer samples a fixed number of member vertices from each hyperedge in the first layer, excluding the parent vertex to avoid immediate backtracking. The third layer further samples new incident hyperedges from these member vertices, excluding the parent hyperedge. Subsequent layers continue to expand alternately between vertices and hyperedges. When $c\in\mathcal{E}$, this procedure is dual to the case of $c\in\mathcal{V}$: we first sample member vertices of the hyperedge, then expand from these members to other incident hyperedges, and so on. For any parent node, when the number of expandable objects is smaller than the sampling budget, we use \texttt{[V-PAD]} for vertices and \texttt{[E-PAD]} for hyperedges to pad the corresponding slots to the fixed size.
In this way, each hypergraph is organized into an alternating incidence tree with a fixed topology but sample-dependent contents. By performing a level-order traversal over this tree, we obtain the detail sequence:
$
    \Pi_{\mathrm{HIDT}}(c)=(u_1,\ldots,u_{L_D})
$,
where each position corresponds to a deterministic structural role. Since the template topology is shared across all samples once the sampling budgets are specified, we can precompute a set of template-level Laplacian positional encodings for the template, such that identical relative structural roles share consistent positional semantics across different samples.

\subsubsection{Order-aware structural overview suffix}
\label{sec:o_suffix}

In order to cover a broader high-order receptive field, we append an order-aware structural overview suffix $\Pi_{\mathrm{O}}(c)$ after HIDT, and refer to the resulting sequence as HIDT-O:
$
    \Pi(c)=\Pi_{\mathrm{HIDT}}(c)\,\Vert\,\Pi_{\mathrm{O}}(c)
$.

The overview suffix no longer directly enumerates additional concrete members, since doing so would lead to a combinatorial explosion in sequence length. 
Instead, it provides a compressed summary for the center object, characterizing how hyperedges with different hop distances and different hyperedge degrees are distributed around it.
Specifically, starting from $c$, we perform a restricted breadth first search (BFS) on the incidence bipartite graph and collect the set of hyperedges $S_h(c)$ that lie strictly in the $h$-th hyperedge layer. Here, $h$ indexes hyperedge layers in the alternating BFS traversal, rather than raw bipartite-edge distance; the center object is treated as the starting layer and only the additionally reached hyperedges are summarized in the overview suffix.
We then partition them according to order buckets:
\begin{equation}
   S_{h,b}(c)=\{e\in S_h(c)\mid \beta(r(e))=b\}, 
\end{equation}
where $\beta(\cdot)$ denotes the order bucket mapping function.
For semantic construction, we adopt a parameter-free alternating aggregation scheme. Let the initial states of vertices and hyperedges be $m_v^{(0)}=\psi(v)$ and $m_e^{(0)}=\psi(e)$, respectively, where $\psi(\cdot)$ denotes the textual embedding of a vertex or a hyperedge. 
We then perform the following alternating propagation on the incidence bipartite graph for $t=1,\ldots,H$, where $t$ denotes the propagation step:
\begin{equation}
\begin{aligned}
    m_e^{(t)}
    &= \frac{1}{|e|}\sum_{v\in e} m_v^{(t-1)}, \\
    m_v^{(t)}
    &= \frac{1}{|\Gamma(v)|}\sum_{e\in\Gamma(v)}
    \Big(m_e^{(t)}+\mathbf{o}_{\beta(r(e))}\Big).
\end{aligned}
\end{equation}
where $\Gamma(v)=\{e\in\mathcal{E}\mid v\in e\}$ denotes the set of hyperedges incident to vertex $v$.
Here, $\mathbf{o}_{\beta(r(e))}$ is a fixed vector assigned to the corresponding order bucket, which is used to explicitly preserve the information of which hyperedge degree interval each hyperedge belongs to during propagation.


It is worth noting that, for each overview slot $(h,b)$, the semantic aggregation depth is tied to its structural hop index. Rather than first propagating to a fixed maximum depth and then using the same final representations for all overview slots, we construct the slot at the $h$-th hop by using the hyperedge states at propagation step $t=h$:
\begin{equation}
\hat m_{h,b}(c)=\mathrm{Mean}\big(\{m_e^{(h)}\mid e\in S_{h,b}(c)\}\big).
\end{equation}
When $S_{h,b}(c)$ is empty, the corresponding overview semantic vector is set to zero.
This design keeps the hop index $h$ consistent across three aspects: the structural layer summarized by $S_{h,b}(c)$, the semantic aggregation depth of $m_e^{(h)}$, and the positional encoding of the overview token.

Overall, HIDT preserves fine-grained local details of the hypergraph, while the overview suffix further supplements high-order structural distributions and cross-hop context across different hop distances and order buckets, without introducing additional parameters.

\subsubsection{Hypergraph token encapsulation}

In HIDT-O, for all vertices or hyperedges involved in sampling, we use off-the-shelf text encoders, such as the Qwen embedding model and SBERT, to encode their textual information. After obtaining HIDT-O, we represent each token $u_i$ as the concatenation of a semantic vector and a structural vector.

On the semantic side, if $u_i$ is a vertex token, we use the embedding of its textual attribute. If $u_i$ is a hyperedge token, we preferentially use the textual embedding of the hyperedge, and fall back to the average of its member vertex representations when no explicit hyperedge text is available. If $u_i$ is an overview token, we use the corresponding overview vector $\hat m_{h,b}$. For pad tokens, we use a zero vector. We denote this semantic representation as $a_i$.
On the structural side, we explicitly maintain a set of structural descriptors for each token:
$
s_i=\big[\,U_i\;\Vert\;\mathbf{t}_{\tau_i}\;\Vert\;\mathbf{h}_{\ell_i}\;\Vert\;\mathbf{o}_{b_i}\;\Vert\;\mathbf{d}_{\delta_i}\,\big]
$,
where $U_i$ denotes the positional encoding, $\mathbf{t}_{\tau_i}$ denotes the token type encoding, $\mathbf{h}_{\ell_i}$ denotes the depth encoding, $\mathbf{o}_{b_i}$ denotes the order bucket encoding over hyperedge degrees, and $\mathbf{d}_{\delta_i}$ denotes the vertex degree bucket encoding. Here, $\delta_i$ denotes the vertex-degree bucket index of the $i$-th token. For token types where a structural descriptor is not applicable, we use a special null bucket.
Finally, the input to the projector is written as:
$g_i=[a_i\,\Vert\,s_i]$.
Through this process, we explicitly provide the structural information that is critical for high-order relations to the projector, rather than leaving it entirely to language modeling for implicit recovery.

\subsection{Hypergraph Incidence Projector}

\subsubsection{Semantic-structural decoupling}

The Hypergraph Incidence Projector (HIP) first maps the semantic vector and the structural vector separately. Given the semantic vector $a_i$ and structural vector $s_i$ of the $i$-th token, it compresses the semantic side into a semantic core:
$
c_i = W_{\mathrm{sem}}\cdot \mathrm{LN}(a_i)
$,
where $\mathrm{LN}(\cdot)$ denotes Layer Normalization. Then, the structural side is mapped into the structural patch space through a role-conditioned structural stem:
$
p_i = W_{\mathrm{str}}^{\rho_i}\cdot \mathrm{LN}(s_i)
$,
where $\rho_i\in\{\mathrm{V},\mathrm{E},\mathrm{O},\mathrm{P}\}$ denotes the four roles of vertex, hyperedge, overview, and pad, respectively.
The two parts are then concatenated and normalized to obtain the initial hidden state of the projector:
$
h_i^{(0)}=\mathrm{LN}\big([c_i\Vert p_i]\big)
$.
In addition, this step decouples the original text embedding dimension from the working width of the projector. The dimension of $h_i^{(0)}$ is usually much smaller than the original text embedding dimension, which avoids linearly increasing the projector size as the text encoder changes.

\subsubsection{Incidence-driven bidirectional vertex-hyperedge message passing}

If the projector only performs independent linear mapping for each token, the high-order structure in the hypergraph is still not truly used. Therefore, HIP introduces a lightweight Hyper-Incidence Block on the hidden states, explicitly performing vertex-hyperedge bidirectional message passing driven by incidences inside the HIP.

Let the initial projector states be $H^{(0)}=\{h_i^{(0)}\}_{i=1}^{L_\mathcal{H}}$. For any detail hyperedge token $e$, let $M(e)$ denote the set of member vertex tokens corresponding to it in the current local sequence. For any detail vertex token $v$, let $N(v)$ denote the set of hyperedge tokens associated with it in the current local sequence. First, in the vertex $\rightarrow$ hyperedge direction, each hyperedge token aggregates messages from its member vertex tokens. We use set attention to model the importance differences among the members inside a hyperedge. For any hyperedge token $e$, the attention weights are defined as:
\begin{equation}
    \alpha_{e\leftarrow v}
=
\mathrm{softmax}_{v\in M(e)}
\left(
\frac{(W_q h_e^{(0)})^\top (W_k h_v^{(0)})}{\sqrt{d_{\mathrm{att}}}}
\right),
\end{equation}
where $d_{\mathrm{att}}$ denotes the attention dimension. The aggregated message from vertices to the hyperedge is obtained as:
\begin{equation}
\vspace{-2pt}
    m_e^{V\rightarrow E}
=
\sum_{v\in M(e)}
\alpha_{e\leftarrow v}\, W_{e\leftarrow v} h_v^{(0)}.
\vspace{-2pt}
\end{equation}
Then, we fuse this message with the current hyperedge state and update the hyperedge representation:
\begin{equation}
    \tilde h_e
=
\mathrm{LN}
\Big(
h_e^{(0)}
+
\phi_E\big([\,h_e^{(0)} \Vert m_e^{V\rightarrow E}\,]\big)
\Big).
\end{equation}
After completing the hyperedge update, we perform reverse aggregation in the hyperedge $\rightarrow$ vertex direction symmetrically. For any vertex token $v$, its associated hyperedge set is $N(v)$, and the corresponding attention weights are:
\begin{equation}
\vspace{-4pt}
    \alpha_{v\leftarrow e}
=
\mathrm{softmax}_{e\in N(v)}
\left(
\frac{(W_q h_v^{(0)})^\top (W_k \tilde h_e)}{\sqrt{d_\mathrm{att}}}
\right).
\end{equation}
The aggregated message from hyperedges to the vertex is written as:
\begin{equation}
\vspace{4pt}
    m_v^{E\rightarrow V}
=
\sum_{e\in N(v)}
\alpha_{v\leftarrow e}\, W_{v\leftarrow e} \tilde h_e,
\end{equation}
and the vertex representation is further updated as:
\begin{equation}
    h_v^{(1)}
=
\mathrm{LN}
\Big(
h_v^{(0)}
+
\phi_V\big([\,h_v^{(0)} \Vert m_v^{E\rightarrow V}\,]\big)
\Big).
\end{equation}
For hyperedge tokens, the final state after this single Hyper-Incidence Block can be written as $h_e^{(1)}=\tilde h_e$.
For tokens that do not participate in the incidence update, such as overview and pad tokens, their states are directly carried over as $h_i^{(1)}=h_i^{(0)}$. After the Hyper-Incidence Block, HIP uses a two-layer MLP to map the final hidden states into the word embedding space of the base LLM,
thereby obtaining the complete hypergraph token sequence
$
T(c)=\{t_i\}_{i=1}^{L_\mathcal{H}}
$.

\subsection{Hypergraph-as-Language Protocol}

Through HIDT-O and HIP, we obtain a structure-aware hypergraph token sequence $T(c)\in\mathbb{R}^{L_\mathcal{H}\times d_{\mathrm{llm}}}$. At the paper level, \emph{Hypergraph as Language} denotes the overall perspective of making hypergraphs directly consumable by LLMs. In this section, we define a Hypergraph-as-Language protocol to refer to its concrete input interface: $T(c)$ is jointly injected into the input sequence of the LLM together with a natural language prompt designed around it, enabling the LLM to complete downstream prediction under a unified question-answering paradigm.

Specifically, for each sample, Hyper-Align uniformly constructs the following three-part prompt:
\begin{equation}
    \text{Prompt} = \text{Background}\;\Vert\;\text{Details}\;\Vert\;\text{Question}.
\end{equation}
The Background provides the task statement and domain description, and embeds the special placeholder \texttt{<hypergraph>} inside the task statement sentence, which is replaced by the hypergraph token sequence $T(c)$ when fed into the LLM. The Details section is a textual context deterministically rendered from the same HIDT-O structure, providing auxiliary natural language supplements for the hypergraph tokens. The Question section specifies the object to be predicted, the candidate label set, and the output format requirements.

\subsection{Hypergraph Alignment Tuning}
During hypergraph alignment tuning, each sample is converted into a dialogue pair. The human side contains the three-part prompt $\mathcal{P}(c)$ defined in Sec.~III-D, where the \texttt{<hypergraph>} placeholder is replaced by the hypergraph token sequence $\mathbf{T}(c)$, while the assistant side contains the target answer text. Let $\mathcal{I}(c)$ denote the resulting input sequence and $y_{1:K}$ denote the target response. Hyper-Align is optimized using the standard causal language modeling loss:
\begin{equation}
    \mathcal{L}_{\mathrm{lm}}
    =
    -\sum_{k=1}^{K}
    \log p_{\Theta,\phi}
    \big(
    y_k \mid y_{<k}, \mathcal{I}(c)
    \big),
\end{equation}
where $\Theta$ denotes the frozen LLM parameters and $\phi$ denotes the trainable HIP parameters. During training, the prompt and inserted hypergraph token regions are masked from supervision, and the language modeling loss is computed only over the response tokens. Hyper-Align therefore follows a projector-only tuning scheme: the base LLM remains fixed, while HIP and its auxiliary prediction heads are jointly optimized to align the hypergraph structure with the LLM token space.

In addition to the language modeling loss, we introduce two auxiliary objectives on the HIP representations. These objectives serve as lightweight regularizers that preserve hyperedge-order information and local incidence relations after semantic-structural fusion and incidence propagation.

The first objective is the order bucket reconstruction loss. Let $\mathcal{I}_{E}$ denote the set of non-padding hyperedge-token positions in the HIDT detail sequence, and let $e_i$ denote the hyperedge represented at position $i$. The hidden state of each such token is used to recover the order bucket of its corresponding hyperedge:
\begin{equation}
    \mathcal{L}_{\mathrm{ord}}
    =
    \sum_{i\in\mathcal{I}_{E}}
    \mathrm{CE}\!\left(
        g_{\mathrm{ord}}\!\left(h_i^{(1)}\right),
        \beta\!\left(r(e_i)\right)
    \right),
\end{equation}
where $g_{\mathrm{ord}}$ denotes the order-bucket prediction head and $\mathrm{CE}$ denotes the cross-entropy loss. Overview, vertex, and padding tokens are excluded from this objective.

The second objective is the relation reconstruction loss. Let $\mathcal{I}_{D}$ denote the set of non-padding positions in the HIDT detail sequence. We sample a set of token pairs $\Omega\subseteq\mathcal{I}_{D}\times\mathcal{I}_{D}$ and predict their local structural relations:
\begin{equation}
    r_{ij}\in
    \{
    \text{unrelated},
    \text{incidence},
    \text{co-member}
    \}.
\end{equation}
Here, ``incidence'' denotes a parent-child vertex-hyperedge pair in the HIDT incidence tree, ``co-member'' denotes two vertex tokens sampled as members of the same parent hyperedge, and ``unrelated'' denotes a sampled pair satisfying neither relation within the HIDT detail sequence. The corresponding loss is
\begin{equation}
    \mathcal{L}_{\mathrm{rel}}
    =
    \sum_{(i,j)\in\Omega}
    \mathrm{CE}\!\left(
        g_{\mathrm{rel}}
        \!\left(
        [h_i^{(1)}\Vert h_j^{(1)}]
        \right),
        r_{ij}
    \right),
\end{equation}
where $g_{\mathrm{rel}}$ denotes the three-way relation prediction head.

The overall training objective is given by:
\begin{equation}
    \mathcal{L}
    =
    \mathcal{L}_{\mathrm{lm}}
    +
    \lambda_{\mathrm{ord}}\mathcal{L}_{\mathrm{ord}}
    +
    \lambda_{\mathrm{rel}}\mathcal{L}_{\mathrm{rel}}.
\end{equation}

\begin{table*}[t]
\centering
\small
\caption{Statistics of the five benchmark datasets in HyperAlign-Bench. All datasets support both vertex classification (VC) and hyperedge classification (HEC).}
\label{tab:Hyper-Align_bench_stats}
\setlength{\tabcolsep}{8pt}
\renewcommand{\arraystretch}{1}
\resizebox{\linewidth}{!}{
\begin{tabular}{l l r r r r c}
\toprule
\textbf{Dataset} & \textbf{Domain / Source} & \textbf{\#Vertices} & \textbf{\#Hyperedges} & \textbf{\#Incidences} & \textbf{\#Classes} & \textbf{Tasks} \\
\midrule
Arxiv-HG & OGBN-Arxiv co-citation hypergraph & 169,343 & 123,826 & 1,116,231 & 40 & VC, HEC \\
Cora-CC  & Cora one-hop successor hypergraph         & 2,341   & 2,219   & 8,162     & 7  & VC, HEC \\
PubMed   & PubMed one-hop successor hypergraph       & 19,716  & 13,011  & 81,502    & 3  & VC, HEC \\
DBLP     & DBLP-A one-hop successor hypergraph       & 2,591   & 2,463   & 4,199     & 6  & VC, HEC \\
IMDB     & IMDB one-hop successor hypergraph         & 3,939   & 839     & 4,656     & 3  & VC, HEC \\
\bottomrule
\end{tabular}
}
\vspace{-12pt}
\end{table*}
\begin{table}[t]
\centering
\small
\caption{Task splits of HyperAlign-Bench for vertex classification (VC) and hyperedge classification (HEC).}
\label{tab:Hyper-Align_bench_splits}
\setlength{\tabcolsep}{14pt}
\renewcommand{\arraystretch}{0.9}
\resizebox{\linewidth}{!}{
\begin{tabular}{l c r r r}
\toprule
\textbf{Dataset} & \textbf{Task} & \textbf{Train} & \textbf{Valid} & \textbf{Test} \\
\midrule
\multirow{2}{*}{Arxiv-HG}
& VC  & 90,941 & 29,799 & 48,603 \\
& HEC & 56,651 & 23,851 & 43,324 \\
\midrule
\multirow{2}{*}{Cora-CC}
& VC  & 1,404 & 468 & 469 \\
& HEC & 1,327 & 451 & 441 \\
\midrule
\multirow{2}{*}{PubMed}
& VC  & 11,829 & 3,943 & 3,944 \\
& HEC & 7,839  & 2,578 & 2,594 \\
\midrule
\multirow{2}{*}{DBLP}
& VC  & 1,554 & 518 & 519 \\
& HEC & 1,485 & 489 & 489 \\
\midrule
\multirow{2}{*}{IMDB}
& VC  & 2,363 & 787 & 789 \\
& HEC & 480   & 163 & 196 \\
\bottomrule
\end{tabular}}
\vspace{-12pt}
\end{table}

\section{HyperAlign-Bench}

To systematically evaluate the ability of different models to capture high-order association structures, we construct HyperAlign-Bench, the first benchmark for hypergraph-language alignment. Unlike existing evaluations that mainly focus on ordinary graphs, HyperAlign-Bench directly treats hypergraphs as the basic data object, preserving the high-order association semantics expressed by vertex-hyperedge incidence. It also provides a unified data construction pipeline, task protocol, and evaluation interface for fair comparison across different methods.

HyperAlign-Bench contains two dual tasks: vertex classification (VC) and hyperedge classification (HEC). VC takes a queried vertex $v\in V$ as the center and predicts its category from its textual attribute $x_v$ together with its query-centered incidence context. Symmetrically, HEC takes a queried hyperedge $e\in E$ as the center and predicts its category from its textual attribute $z_e$ and the corresponding incidence context. From the incidence-dual perspective, HEC can be viewed as vertex classification on the dual hypergraph, where the original hyperedges become the queried objects. Therefore, the two tasks follow the same object-centric protocol, enabling evaluation from both sides of the vertex-hyperedge incidence relation.

In HyperAlign-Bench, the main dataset is Arxiv-HG, which is built upon OGBN-Arxiv~\cite{ogbn}. We convert paper citation relations into a co-citation hypergraph: each paper is represented as a vertex with its title and abstract as textual attributes, while for each source paper, the set of papers it cites is regarded as a hyperedge. The textual information and subject category of the source paper are respectively associated with the induced hyperedge as its textual attribute and category. Thus, VC and HEC provide two symmetric classification views of the same incidence structure, with the queried object's own textual attribute available in both tasks. This construction avoids flattening high-order co-citation relations into ordinary graph edges, resulting in a training and in-domain evaluation dataset with 169,343 vertices and 123,826 hyperedges. 
In addition to the main dataset, HyperAlign-Bench includes 4 reorganized datasets derived from HyperBERT~\cite{hyperbert}, namely Cora-CC, PubMed, DBLP, and IMDB, covering different domains such as papers, author relations, and movies. These datasets are used to evaluate the model's ability to transfer high-order association structure modeling to unseen domains. Overall, HyperAlign-Bench provides an experimental foundation for validating hypergraph-native alignment capability and cross-domain generalization. The basic statistics of HyperAlign-Bench are summarized in Tab.~\ref{tab:Hyper-Align_bench_stats}, and the train, validation, and test splits are reported in Tab.~\ref{tab:Hyper-Align_bench_splits}.

\begin{figure}[t]
    \centering
    \includegraphics[width=\linewidth]{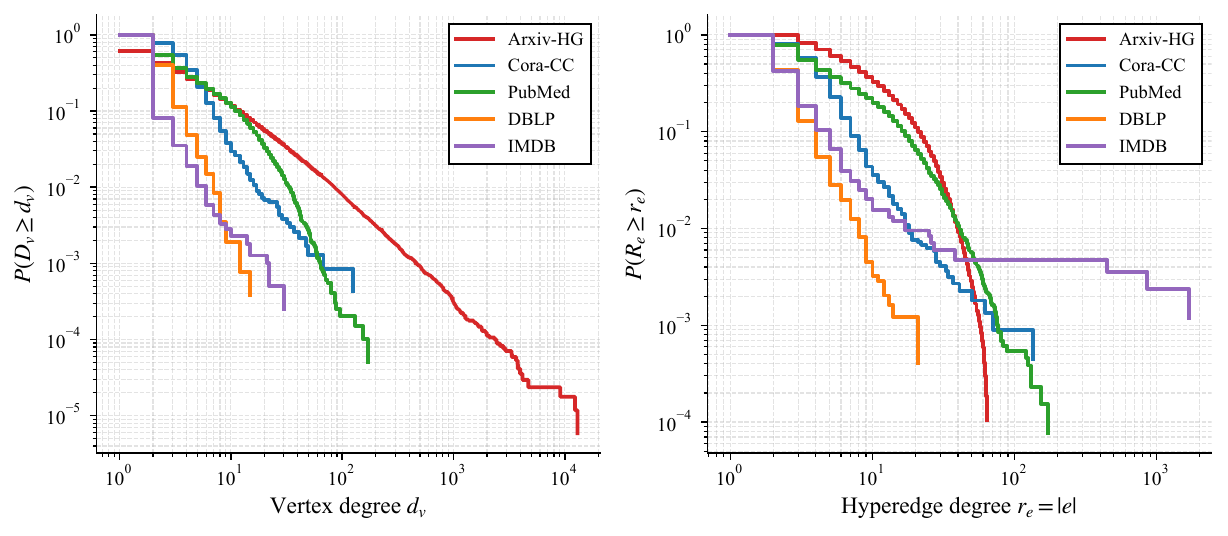}
    \vspace{-12pt}
    \caption{CCDFs of vertex degree and hyperedge degree on the five benchmark datasets in HyperAlign-Bench. Vertex degree denotes the number of incident hyperedges of a vertex, while hyperedge degree denotes the number of vertices contained in a hyperedge.}
    \label{fig:bench_degree_ccdf}
    \vspace{-12pt}
\end{figure}

We further visualize the complementary cumulative distribution functions (CCDFs) of vertex degree and hyperedge degree in Fig.~\ref{fig:bench_degree_ccdf}. Here, vertex degree denotes the number of incident hyperedges of a vertex, while hyperedge degree denotes the number of vertices contained in a hyperedge. The distributions show that HyperAlign-Bench covers diverse high-order structural patterns across datasets. Arxiv-HG has the largest scale and exhibits a long-tailed vertex-degree distribution, reflecting the highly skewed connectivity patterns in citation-derived hypergraphs. PubMed and Cora-CC also contain non-trivial high-order structures, while DBLP is relatively more concentrated. IMDB contains fewer hyperedges, but its hyperedge-degree distribution is highly long-tailed, indicating the presence of very large hyperedges. These differences make HyperAlign-Bench suitable for evaluating whether a model can generalize across heterogeneous hypergraph structures rather than overfitting to a single degree regime.

\section{Experimental Results}

\subsection{Experimental Setup}

We evaluate Hyper-Align on HyperAlign-Bench through both in-domain and cross-domain settings. The model is jointly trained for vertex classification (VC) and hyperedge classification (HEC) on Arxiv-HG, where both tasks use the 40 computer science categories of OGBN-Arxiv. Cross-domain zero-shot evaluation is conducted on Cora-CC, PubMed, DBLP, and IMDB. For each task, the corresponding checkpoint trained on Arxiv-HG is directly transferred to the target datasets without additional fine-tuning.



\begin{table}[t]
\centering
\small
\caption{In-domain performance on VC and HEC tasks. Here, ``GOFA$^\ast$'' indicates that the evaluation is conducted using the officially released weights directly.}
\vspace{-6pt}
\label{tab:performance}
\setlength{\tabcolsep}{3pt} 
\renewcommand{\arraystretch}{1} 
\resizebox{\linewidth}{!}{
\begin{tabular}{l l l c c}
\toprule
\textbf{Category} & \textbf{Method} & \textbf{Venue} & \textbf{VC (\%)} & \textbf{HEC (\%)} \\
\midrule
\multirow{4}{*}{HGNNs}
& HGNN \cite{hgnn}        & AAAI'19 & 69.5 & 69.4 \\
& HyperGCN \cite{hypergcn}    & NeurIPS'19      & 71.6 & 71.6 \\
& HAN \cite{han}         & WWW'19& 65.3 & 66.5 \\
& AllSetTransformer \cite{allset} & ICLR'22      & 68.9 & 70.0 \\
\midrule
PLM-based
& HyperBERT \cite{hyperbert}   & EMNLP'24      & 59.1 & 58.5 \\
\midrule
\multirow{4}{*}{General LLMs}
& Llama2-7B \cite{llama2}   & arXiv'23      & 9.7  & 8.1  \\
& Llama3-8B \cite{llama3}   & arXiv'24      & 54.5 & 56.0 \\
& Qwen3-8B \cite{qwen3}    & arXiv'25      & 55.5 & 60.3 \\
& GPT-5-mini \cite{gpt5mini}  & OpenAI'25      & 65.4 & 67.0   \\
\midrule
\multirow{8}{*}{Graph-LLMs}& GraphGPT \cite{graphgpt}    & SIGIR'24      & 69.3 & 69.7   \\
& LLaGA \cite{llaga}       & ICML'24      & 70.4 & 71.5 \\
& TEA-GLM \cite{tea_glm}     & NeurIPS'24      & 70.8 & 71.3 \\
& GraphPrompter \cite{graphprompter}   & ICDE'25      & 60.1 & 68.6 \\
& PromptGFM \cite{promptgfm}   & arXiv'25      & 68.9 & 69.7 \\
& UniGraph \cite{unigraph}   & SIGKDD'25      & 62.4 & 71.2 \\
& GOFA$^\ast$ \cite{gofa} & ICLR'25      & 51.1 & 53.1 \\
& GOFA \cite{gofa}        & ICLR'25      & 69.7 & 70.5 \\
\midrule
Hypergraph-LLMs
& \textbf{Hyper-Align (Ours)} & --   & \textbf{76.9} & \textbf{78.2} \\
\bottomrule
\end{tabular}
}
\vspace{-8pt}
\end{table}
\subsubsection{Implementation Details}
By default, Hyper-Align uses Qwen3-8B~\cite{qwen3} as the frozen backbone LLM and Qwen3-Embedding-0.6B~\cite{qwen3_emb} to pre-encode textual semantics. Only HIP, including its auxiliary prediction heads, is optimized. Training is performed on 4 NVIDIA A100 GPUs using DeepSpeed ZeRO-2, bf16 precision, and gradient checkpointing. We jointly train the VC and HEC tasks for 2 epochs with a per-GPU batch size of 8 and 2 gradient accumulation steps, resulting in an effective batch size of 64. The learning rate is set to $2\times10^{-3}$ with cosine decay and a warmup ratio of 0.03.
The maximum sequence length is 4096.

HIDT-O retains at most 160 hypergraph tokens. The HIDT incidence tree has a maximum depth of 3, with branching budgets of 8 incident hyperedges, 8 member vertices per hyperedge, and 1 additional hyperedge at deeper vertex layers. The overview suffix contains 2 hyperedge hops and 4 order buckets. The auxiliary loss weights $\lambda_{\mathrm{ord}}$ and $\lambda_{\mathrm{rel}}$ are both set to 0.01. At inference, we adopt greedy decoding.

\subsubsection{Compared baselines}

We compare against four baseline families. Hypergraph Neural Networks (HGNNs) include the classical HGNN~\cite{hgnn}, HyperGCN~\cite{hypergcn}, HAN~\cite{han}, and AllSetTransformer~\cite{allset}. For PLM-based hypergraph learning method, we select HyperBERT~\cite{hyperbert}, which is the only existing work in this category. General LLMs include Llama2-7B~\cite{llama2}, Llama3-8B~\cite{llama3}, Qwen3-8B~\cite{qwen3}, and GPT-5-mini~\cite{gpt5mini}. Graph-LLMs include GraphGPT~\cite{graphgpt}, LLaGA~\cite{llaga}, TEA-GLM~\cite{tea_glm}, GraphPrompter~\cite{graphprompter}, PromptGFM~\cite{promptgfm}, UniGraph~\cite{unigraph}, and GOFA~\cite{gofa}. 

HGNNs use Qwen3-Embedding-0.6B features encoded from object titles and abstracts, with hyperedge representations obtained through their native vertex-to-hyperedge aggregation; HyperBERT uses the same textual inputs. General LLMs are evaluated zero-shot using only textual prompts and serialized contexts. For graph-LLMs, we apply clique expansion for VC and clique expansion of the dual hypergraph for HEC, where graph nodes represent original hyperedges and edges indicate nonempty overlap. The converted graphs retain the original labels, data splits, and textual features. Cross-domain evaluation directly transfers the corresponding Arxiv-HG-trained model without target-domain fine-tuning. Unless otherwise stated, we use official implementations and recommended settings. $\mathrm{GOFA}^{*}$ denotes direct evaluation using the released weights, whereas GOFA denotes the model retrained under our benchmark protocol.

\subsection{Comparison with Existing Methods}

\subsubsection{In-domain evaluation}
Table~\ref{tab:performance} reports the in-domain results on Arxiv-HG. Hyper-Align achieves the best performance on both VC and HEC, significantly outperforming the strongest baseline. Among existing hypergraph-specific methods, HGNNs achieve competitive results by directly modeling hypergraph structures, but they remain supervised encoders tailored to specific tasks, offering no interface for alignment with language nor supporting any zero-shot capabilities. The pre-trained language model (PLM)-based method HyperBERT performs much worse, showing that simply injecting textual semantics into PLMs is insufficient for hypergraph modeling. General LLMs improve when stronger instruction-following models are used, but their performance is still limited because textual prompts alone cannot faithfully represent native vertex-hyperedge incidence structures. Graph-LLMs further benefit from structure-aware adaptation, yet they are fundamentally built on pairwise graph representations and therefore cannot fully preserve hyperedge-level grouping semantics. In contrast, our Hyper-Align is the first hypergraph-native LLM framework that directly aligns vertex-hyperedge incidence structures with the LLM token space. The substantial improvement over all HGNNs, PLM-based methods, general LLMs, and graph-LLMs demonstrates both the necessity of native hypergraph-language alignment and the effectiveness of our proposed design.

\begin{table*}[t]
\centering
\small
\caption{Zero-shot performance on Cora-CC, PubMed, DBLP and IMDB datasets.}
\vspace{-8pt}
\label{tab:zero_shot_results}
\setlength{\tabcolsep}{10pt}
\renewcommand{\arraystretch}{1}
\resizebox{\linewidth}{!}{
\begin{tabular}{l c cc cc cc cc cc}
\toprule
\multirow{2}{*}{\textbf{Method}} & \multirow{2}{*}{\textbf{Venue}}
& \multicolumn{2}{c}{\textbf{Cora-CC (\%)}}
& \multicolumn{2}{c}{\textbf{PubMed (\%)}}
& \multicolumn{2}{c}{\textbf{DBLP (\%)}}
& \multicolumn{2}{c}{\textbf{IMDB (\%)}}
& \multicolumn{2}{c}{\textbf{Average (\%)}} \\
\cmidrule(lr){3-4}
\cmidrule(lr){5-6}
\cmidrule(lr){7-8}
\cmidrule(lr){9-10}
\cmidrule(lr){11-12}
& & \textbf{VC} & \textbf{HEC}
  & \textbf{VC} & \textbf{HEC}
  & \textbf{VC} & \textbf{HEC}
  & \textbf{VC} & \textbf{HEC}
  & \textbf{VC} & \textbf{HEC} \\
\midrule
GraphGPT \cite{graphgpt}        & SIGIR'24   & 60.3 & 62.3 & 71.2 & 72.8 & 50.0 & 59.8 & 30.3 & 29.6 & 53.0 & 56.1 \\
LLaGA \cite{llaga}              & ICML'24    & 2.8  & 3.6  & 0.9  & 0.5  & 51.1 & 56.4 & 18.1 & 28.1 & 18.2 & 22.2 \\
TEA-GLM \cite{tea_glm}          & NeurIPS'24 & 32.6 & 51.2 & 12.5 & 26.3 & 58.2 & 60.5 & 52.3 & 32.7 & 38.9 & 42.7 \\
Graph Prompter \cite{graphprompter} & ICDE'25    & 32.8 & 37.6 & 58.0 & 60.4 & 61.1 & 60.3 & 52.3 & 41.3 & 51.1 & 49.9 \\
PromptGFM \cite{promptgfm}      & arXiv'25   & 58.4 & 55.1 & 70.8 & \underline{75.9} & 56.3 & 60.9 & 28.9 & 27.0 & 53.6 & 54.7 \\
UniGraph \cite{unigraph}        & SIGKDD'25  & \underline{64.4} & \underline{72.5} & \underline{72.0} & 66.5 & \underline{65.5} & \underline{61.5} & 51.2 & 42.3 & \underline{63.3} & \underline{60.7} \\
GOFA$^\ast$ \cite{gofa}         & ICLR'25    & 61.4 & 60.8 & 70.2 & 71.1 & 4.4 & 2.1 & \underline{63.6} & 28.6 & 49.9 & 40.7 \\
GOFA \cite{gofa}                & ICLR'25    & 59.3 & 60.8 & 69.9 & 71.3 & 3.1 & 2.1 & 54.0 & \underline{42.4} & 46.6 & 44.2 \\
\midrule
\textbf{Hyper-Align (Ours)}     & --         & \textbf{74.8} & \textbf{75.7} & \textbf{77.5} & \textbf{77.6} & \textbf{67.2} & \textbf{64.6} & \textbf{74.5} & \textbf{44.9} & \textbf{73.5} & \textbf{65.7} \\
\bottomrule
\end{tabular}
}
\vspace{-8pt}
\end{table*}


\subsubsection{Cross-domain zero-shot evaluation}
Table~\ref{tab:zero_shot_results} further reports zero-shot results on four unseen datasets. Hyper-Align obtains the best performance on VC and HEC tasks across all datasets. Although several graph-LLMs achieve competitive scores on individual datasets, their performance varies substantially across domains. In contrast, Hyper-Align shows more stable cross-domain generalization, indicating that Hyper-Align learns a more transferable alignment pattern for high-order association modeling rather than fitting only the source dataset.

\vspace{-8pt}

\subsection{Ablation Study}

\begin{table*}[t]

\centering
\small
\caption{Detailed ablation study of the proposed components.}
\vspace{-8pt}
\label{tab:ablation_settings}
\setlength{\tabcolsep}{8pt}
\renewcommand{\arraystretch}{1}
\resizebox{\linewidth}{!}{
\begin{tabular}{lccccccccccc}
\toprule
\multirow{2}{*}{\textbf{Exp.}} 
& \multicolumn{2}{c}{\textbf{HIDT-O}} 
& \multirow{2}{*}{\textbf{HIP}} 
& \multicolumn{2}{c}{\textbf{Aux. Loss}} 
& \multicolumn{2}{c}{\textbf{Input Protocol}} 
& \multicolumn{2}{c}{\textbf{In-domain}} 
& \multicolumn{2}{c}{\textbf{Zero-shot Avg}} \\
\cmidrule(lr){2-3}
\cmidrule(lr){5-6}
\cmidrule(lr){7-8}
\cmidrule(lr){9-10}
\cmidrule(lr){11-12}
& \textbf{HIDT} 
& \textbf{Overview} 
& 
& \textbf{OBRecon.} 
& \textbf{RelRecon.} 
& \textbf{HGToken} 
& \textbf{Details} 
& \textbf{VC}& \textbf{HEC}& \textbf{VC}& \textbf{HEC}\\
\midrule
F0: Full Hyper-Align      
& \cmark & \cmark & \cmark & \cmark & \cmark & \cmark & \cmark 
& \textbf{76.9} & \textbf{78.2} & \textbf{73.5} & \textbf{65.7} \\

S1: w/o HIDT detail  
& \xmark   & \cmark & \cmark & \cmark & \cmark & \cmark & \cmark 
& 70.4 & 71.1 & 66.8 & 58.9 \\

S2: w/o Overview     
& \cmark & \xmark & \cmark & \cmark & \cmark & \cmark & \cmark 
& 75.9 & 76.7 & 72.4 & 64.2 \\

P1: MLP projector    
& \cmark & \cmark & \xmark & \cmark & \cmark & \cmark & \cmark 
& 74.3 & 75.0 & 70.9 & 62.1 \\

L1: w/o OBRecon.     
& \cmark & \cmark & \cmark & \xmark & \cmark & \cmark & \cmark 
& 76.5 & 77.5 & 73.0 & 65.0 \\

L2: w/o RelRecon.    
& \cmark & \cmark & \cmark & \cmark & \xmark & \cmark & \cmark 
& 76.4 & 77.6 & 72.9 & 65.1 \\

L3: w/o Aux. losses  
& \cmark & \cmark & \cmark & \xmark & \xmark & \cmark & \cmark 
& 76.1 & 77.3 & 72.5 & 64.4 \\

G1: Text-only prompt 
& --     & --     & --     & --     & --     & \xmark & \cmark 
& 56.5 & 62.0 & 61.3& 58.2\\

G2: HG token only    
& \cmark & \cmark & \cmark & \cmark & \cmark & \cmark & \xmark 
& 75.2 & 76.4 & 70.8 & 62.7 \\
\bottomrule
\end{tabular}
}
\vspace{-12pt}
\end{table*}

\subsubsection{Ablation on the proposed components}
Table~\ref{tab:ablation_settings} presents a detailed ablation study of the proposed components. Removing the HIDT sequence causes the largest degradation. This confirms that fine-grained vertex-hyperedge incidence details are crucial for modeling native high-order associations. Removing the overview suffix also leads to consistent drops on both in-domain and zero-shot evaluation, indicating that overview tokens provide complementary broader structural context beyond the local HIDT template.
For the projector, replacing HIP with a plain MLP projector substantially hurts performance, especially under zero-shot transfer, showing that independent token projection is insufficient for aligning hypergraph structures with the LLM space. The auxiliary losses also bring consistent gains: removing either auxiliary reconstruction task degrades performance, while removing both leads to a larger drop. This suggests that the two auxiliary objectives serve as useful regularizers for preserving order-aware and relation-aware structural information.
We also ablate the Hypergraph-as-Language protocol. The text-only prompt variant, which removes \texttt{<hypergraph>} and relies only on textual Details, performs much worse than full Hyper-Align. In contrast, the hypergraph-token-only variant remains competitive but still underperforms the full protocol, particularly in zero-shot evaluation. These results indicate that the hypergraph tokens carry the main structural information, while textual details further facilitate LLM alignment and cross-domain generalization.

\subsubsection{Effect of base LLM and embedding model}
Table~\ref{tab:base_llm_embedding} evaluates Hyper-Align with different choices of base LLMs and embedding models. This experiment is designed to examine whether the advantage of Hyper-Align comes merely from using stronger recent LLMs or from the proposed hypergraph-native architecture itself.
To this end, we include Vicuna-7B \cite{vicuna} with SBERT \cite{sbert} and SimTeG \cite{simteg} embeddings, which follow the commonly used settings in prior graph-LLM work such as LLaGA~\cite{llaga}; in particular, Vicuna-7B with SimTeG corresponds to the default LLaGA setting. 

\begin{table}[t]

\centering
\small
\caption{Effect of using different base LLMs and embedding models.}
\vspace{-6pt}
\label{tab:base_llm_embedding}
\setlength{\tabcolsep}{3pt}
\renewcommand{\arraystretch}{1}
\resizebox{\linewidth}{!}{
\begin{tabular}{llcccccc}
\toprule
\multirow{2}{*}{\textbf{Base LLM}}
& \multirow{2}{*}{\textbf{Emb. Model}}
& \multirow{2}{*}{\textbf{Emb. Dim.}}
& \multicolumn{2}{c}{\textbf{In-domain}}
& \multicolumn{2}{c}{\textbf{Zero-shot Avg}} \\
\cmidrule(lr){4-5}
\cmidrule(lr){6-7}
& & & \textbf{VC} & \textbf{HEC} & \textbf{VC} & \textbf{HEC} \\
\midrule
Vicuna-7B & SBERT & 384 & 75.8 & 76.5 & 70.9 & 61.2 \\
Vicuna-7B & SimTeG & 2432 & 76.0 & 77.1 & 72.0 & 61.9 \\
Qwen3-8B & Qwen3-Emb-0.6B & 1024 & 76.9 & 78.2 & 73.5 & 65.7 \\
Qwen3-8B & Qwen3-Emb-4B & 2560 & \textbf{77.2} & \textbf{78.8} & \textbf{74.2} & \textbf{66.5} \\
\bottomrule
\end{tabular}
}
\vspace{-12pt}
\end{table}

Under these controlled settings, Hyper-Align still achieves strong performance and remains substantially better than graph-LLM baselines reported in Table~\ref{tab:performance}, demonstrating that the gain is not simply due to using a more advanced LLM or embedding model. Meanwhile, replacing SBERT with SimTeG improves the results under the Vicuna-7B setting, and Qwen3-based configurations \cite{qwen3,qwen3_emb} further strengthen both in-domain and zero-shot performance. These results show that Hyper-Align is compatible with different LLMs and semantic encoders, while its main advantage comes from the hypergraph-native alignment design rather than being tied to a specific LLM or embedding model.

\begin{table}[t]
\centering
\small
\caption{Comparison of single-task and joint training.}
\label{tab:joint_vs_single_task}
\renewcommand{\arraystretch}{1}
\setlength{\tabcolsep}{16pt}
\vspace{-6pt}
\resizebox{\linewidth}{!}{
\begin{tabular}{lcccc}
\toprule
\multirow{2}{*}{\textbf{Training}}
& \multicolumn{2}{c}{\textbf{In-domain}}
& \multicolumn{2}{c}{\textbf{Zero-shot Avg}}
\\
\cmidrule(lr){2-3}
\cmidrule(lr){4-5}
& \textbf{VC} & \textbf{HEC} & \textbf{VC} & \textbf{HEC} \\
\midrule
Single-task & \textbf{77.0} & 77.6 & 67.3 & 64.7 \\
Joint        & 76.9 & \textbf{78.2} & \textbf{73.5} & \textbf{65.7} \\
\bottomrule
\end{tabular}
}
\vspace{-8pt}
\end{table}


\subsubsection{Single-task \& joint training}
Table~\ref{tab:joint_vs_single_task} compares single-task training with joint training. In-domain performance is roughly the same for both training methods, but joint training has a significant advantage in cross-domain zero-shot performance. These results indicate that jointly optimizing vertex-centered and hyperedge-centered tasks encourages HIP to learn a more general shared alignment between hypergraphs and language, resulting in better generalization ability.

\begin{table}[t]
\centering
\caption{Model performance of setting different training epochs.}
\label{tab:joint_epoch_scaling}
\setlength{\tabcolsep}{16pt}
\resizebox{\linewidth}{!}{
\begin{tabular}{ccccc}
\toprule
\multirow{2}{*}{\textbf{Epoch}}
& \multicolumn{2}{c}{\textbf{In-domain}}
& \multicolumn{2}{c}{\textbf{Zero-shot Avg}} \\
\cmidrule(lr){2-3}
\cmidrule(lr){4-5}
& \textbf{VC} & \textbf{HEC} & \textbf{VC} & \textbf{HEC} \\
\midrule
1 & 76.6 & 78.0 & 68.3 & 61.6 \\
2 &76.9 & 78.2 & 73.5 & 65.7 \\
3 & 77.0 & 78.4 & 63.1 & 58.8 \\
\bottomrule
\end{tabular}}
\end{table}

\subsubsection{Effect of training epochs}
Table~\ref{tab:joint_epoch_scaling} studies the effect of training epochs. Increasing the training length from one epoch to two epochs improves zero-shot transfer substantially, indicating that sufficient projector tuning is necessary for aligning high-order structures with the frozen LLM. A third epoch gives only marginal in-domain improvement, but clearly hurts zero-shot performance. This suggests that excessive tuning may overfit the source Arxiv-HG distribution and weaken cross-domain generalization. Therefore, we use two training epochs as the default setting.


\begin{figure}[t]
\vspace{-10pt}
    \centering
    \includegraphics[width=1\linewidth]{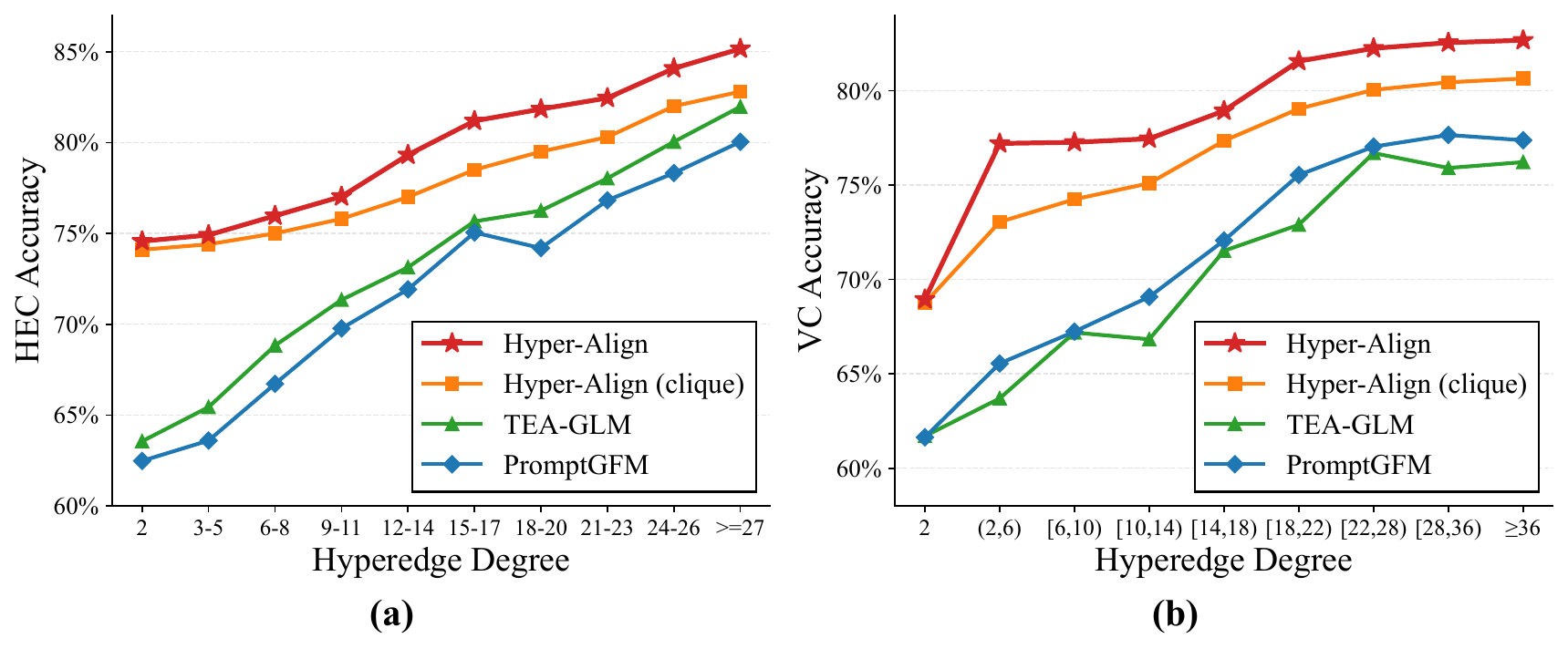}
    \vspace{-20pt}
    \caption{Dual structural stratification analysis on Arxiv-HG. 
    Left: HEC performance under different hyperedge-degree ranges. 
    Right: VC performance under different ranges of the average degree of hyperedges incident to the queried vertex.}
    \label{fig:dual_degree_analysis}
    \vspace{-12pt}
\end{figure}


\subsubsection{Structural analysis across hyperedge degrees}

We further analyze Hyper-Align from two complementary structural views on Arxiv-HG. 
Fig.~\ref{fig:dual_degree_analysis}(a) stratifies HEC test hyperedges by their hyperedge degree, while Fig.~\ref{fig:dual_degree_analysis}(b) stratifies VC test vertices by the average degree of their incident hyperedges. 
The former provides a hyperedge-centered view, measuring how models behave when the target hyperedge itself contains different numbers of vertices. 
The latter provides a vertex-centered dual view, measuring how vertex classification changes when the queried vertex is surrounded by hyperedges with different average degrees. 
Together, these two views characterize whether a model can benefit from increasingly rich high-order associations.
We compare our Hyper-Aligh to representative graph-LLMs, \ie, TEA-GLM and PromptGFM. We also additionally include Hyper-Align (Clique), an internal pairwise variant that replaces each native hyperedge with its clique-expanded pairwise edges while keeping the remaining framework unchanged.

From the hyperedge-centered view, Hyper-Align consistently achieves the best performance across all degree ranges. 
When the hyperedge degree is small, especially in the degree-2 case, Hyper-Align already outperforms graph-LLMs, showing that our hypergraph formulation is naturally compatible with pairwise relations and can model them effectively. 
As the hyperedge degree increases, the advantage of Hyper-Align over Hyper-Align (clique) becomes more evident. 
This suggests that clique expansion can capture part of the relational signal, but cannot adequately preserve the native grouping semantics of hyperedges.
A consistent trend also appears from the vertex-centered view. 
When a vertex is mainly connected to low-degree hyperedges, the local structure is closer to an ordinary pairwise or low-order setting, where Hyper-Align already remains competitive. 
As the average degree of incident hyperedges increases, Hyper-Align maintains the strongest performance and shows a larger advantage over the clique-based variant. 
This indicates that the benefit of native hypergraph modeling is not limited to classifying hyperedges themselves; it also improves vertex understanding when the surrounding context contains richer group-level associations.

Overall, the two stratified analyses lead to the same conclusion: Hyper-Align is effective in both low-order and high-order regimes, while its advantage becomes more pronounced as local high-order connectivity becomes richer. 
These results further support our claim that preserving vertex-hyperedge incidence structure is crucial for modeling high-order associations, and that reducing hypergraphs to pairwise graphs loses important structural semantics from both the hyperedge-centered and vertex-centered perspectives.

\section{Conclusion}

In this paper, we introduce the \emph{Hypergraph as Language} perspective and propose Hyper-Align, the first hypergraph-native alignment framework for LLMs. Different from existing graph-LLMs that rely on pairwise graph representations, Hyper-Align directly represents vertex-hyperedge incidence structures through a Hypergraph-as-Language protocol, encodes fine-grained high-order contexts with HIDT-O, and aligns hypergraph tokens with the LLM token space via HIP.
To support systematic evaluation, we construct HyperAlign-Bench, which covers both vertex-level and hyperedge-level tasks under in-domain and zero-shot settings. Extensive experiments show that our Hyper-Align substantially outperforms current HGNNs, PLM-based methods, general LLMs, and graph-LLMs. We hope Hyper-Align provides a useful foundation for building language-aligned models that can understand and reason over complex high-order associations.


 
\bibliographystyle{IEEEtran}
\bibliography{a_cite}
%


\section{Biography}
\vspace{-24pt}

\begin{IEEEbiography}[{\includegraphics[width=1in,height=1.25in,clip,keepaspectratio]{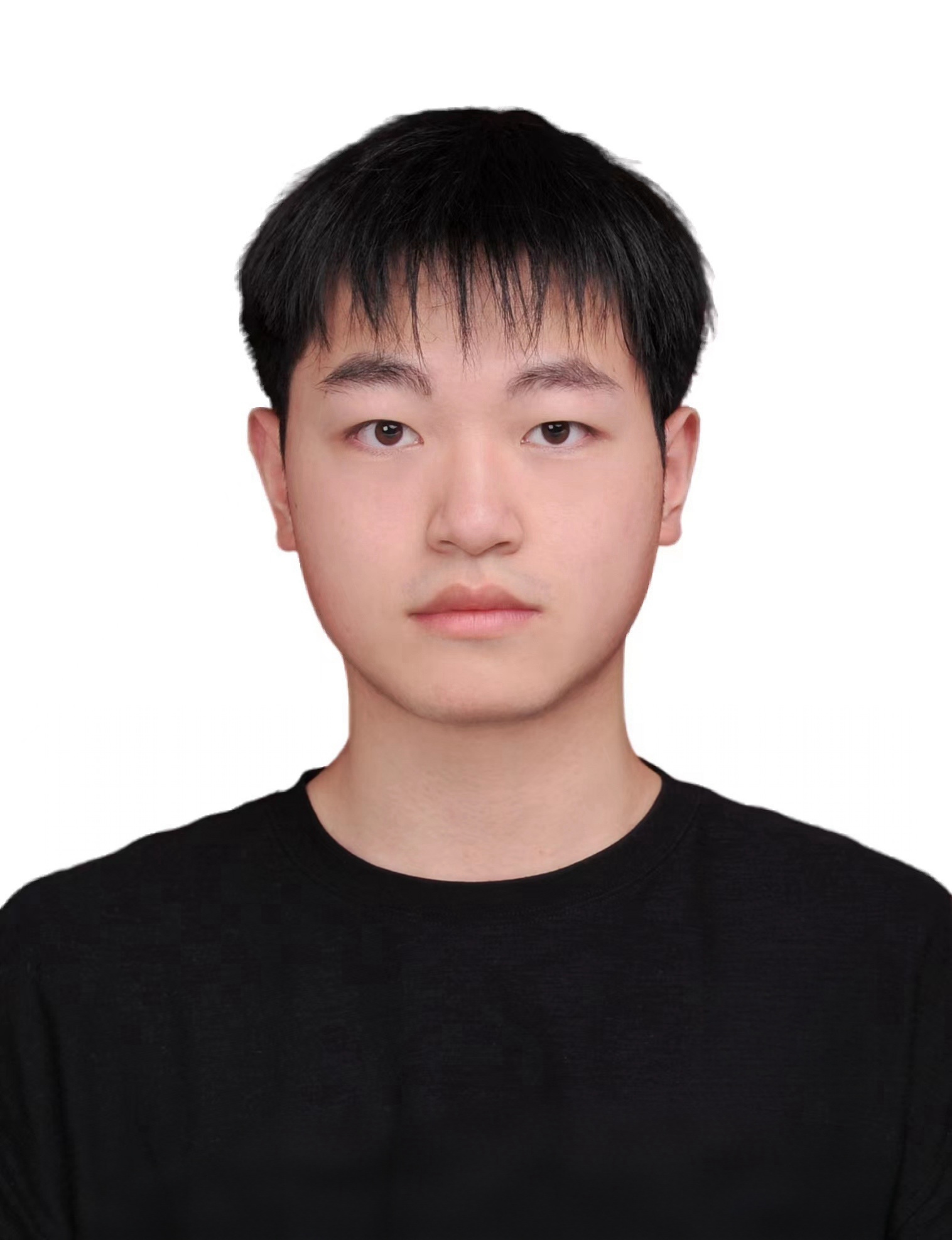}}]{Mengqi Lei} is currently working toward the Ph.D. degree in the School of Software, Tsinghua University, Beijing. He received the B.S. degree from the Department of Computer Science, China University of Geosciences, Wuhan, China. His research interests include hypergraph computation, computer vision, and vision language models.
\end{IEEEbiography}
\vspace{-16pt}

\begin{IEEEbiography}[{\includegraphics[width=1in,height=1.25in,clip,keepaspectratio]{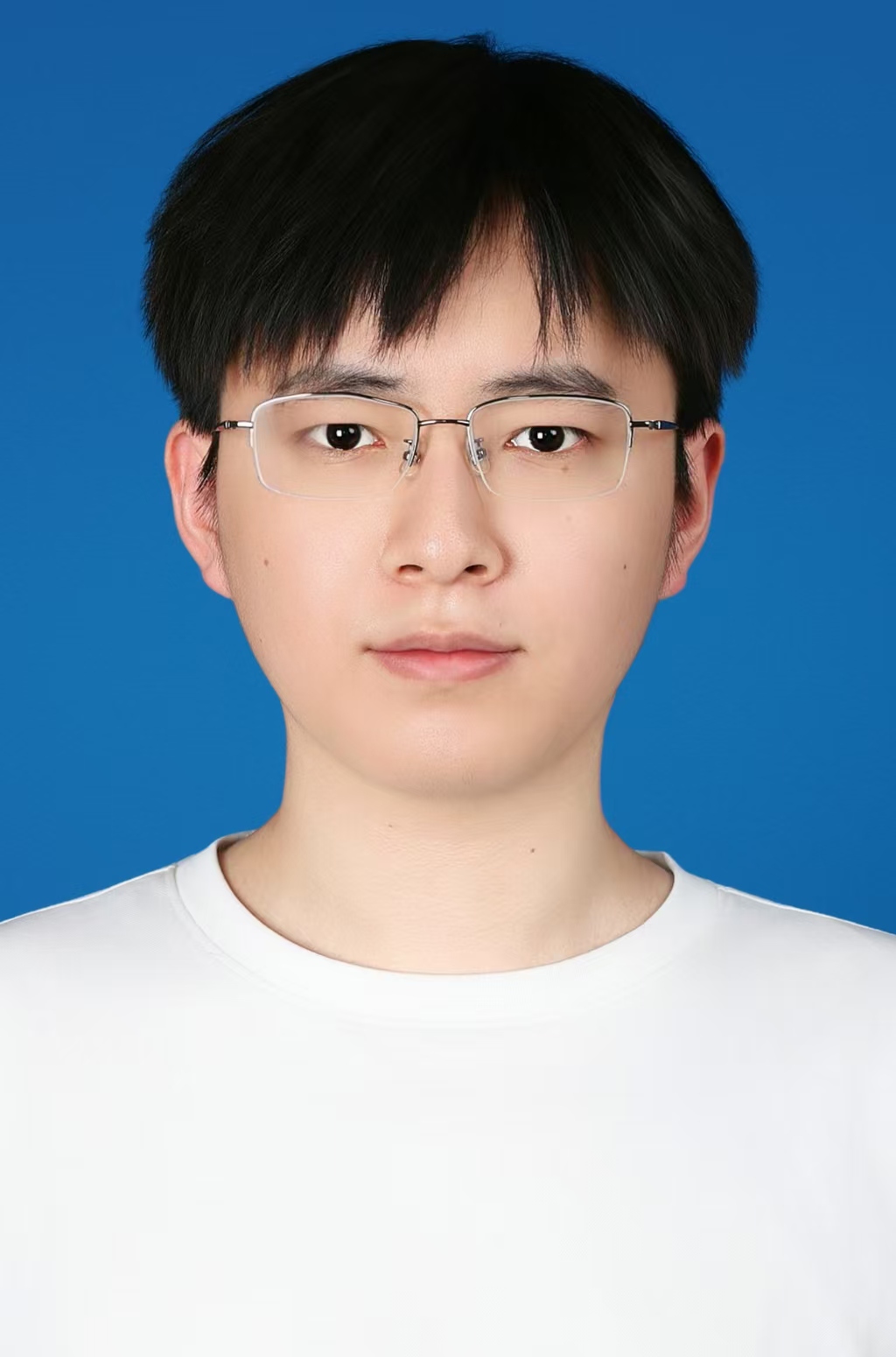}}]{Guohuan Xie} received his B.E. degree from the College of Software, Nankai University, in 2026. He is currently pursuing his master's degree at the School of Software, Tsinghua University. His research interests include computer vision and machine learning.
\end{IEEEbiography}
\vspace{-16pt}

\begin{IEEEbiography}[{\includegraphics[width=1in,height=1.25in,clip,keepaspectratio]{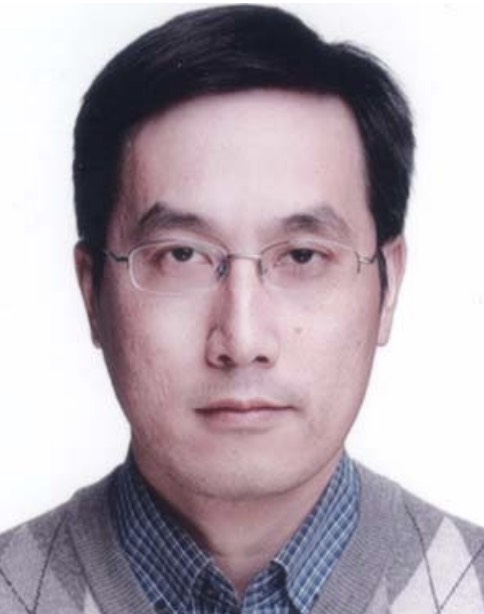}}]{Shihui Ying} received the BEng degree in mechanical engineering and the PhD degree in applied mathematics from Xi’an Jiaotong University, Xi’an, China, in 2001 and 2008, respectively. He is currently a professor with the Department of Mathematics, School of Science, Shanghai University, Shanghai, China. His current research interests include geometric theory and methods for machine intelligence and medical image analysis.
\end{IEEEbiography}
\vspace{-16pt}

\begin{IEEEbiography}[{\includegraphics[width=1in,height=1.25in,clip,keepaspectratio]{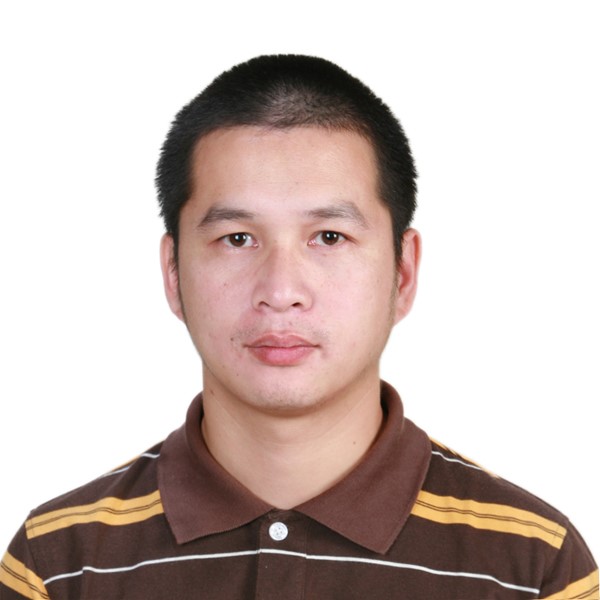}}]{Shaoyi Du} is a professor at Xi’an Jiaotong University. He received double Bachelor degrees in computational mathematics and in computer science in 2002 and received his M.S. degree in applied mathematics in 2005 and Ph.D. degree in pattern recognition and intelligence system from Xi’an Jiaotong University, China in 2009. His research interests include computer vision, machine learning and pattern recognition.
\end{IEEEbiography}

\vspace{-12pt}

\begin{IEEEbiography}[{\includegraphics[width=1in,height=1.25in,clip,keepaspectratio]{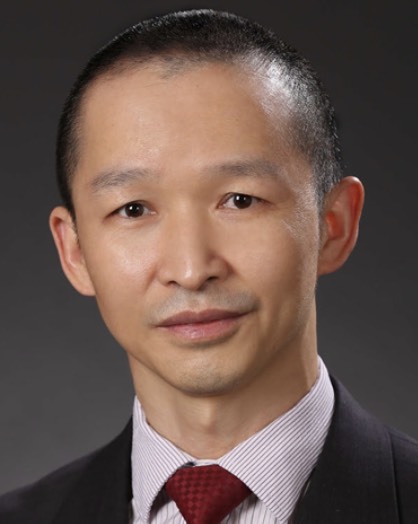}}]{Jun-Hai Yong} received the B.S. and Ph.D. degrees in computer science from Tsinghua University, Beijing, China, in 1996 and 2001, respectively. He held a visiting researcher position with the Department of Computer Science, Hong Kong University of Science and Technology in 2000. He was a PostDoctoral Fellow with the Department of Computer Science, University of Kentucky, from 2000 to 2002. He is currently a Professor with the School of Software, Tsinghua University. His main research interests include computer-aided design and computer graphics. He received a lot of awards, such as the National Excellent Doctoral Dissertation Award, the National Science Fund for Distinguished Young Scholars, the Best Paper Award of the ACM SIGGRAPH Eurographics Symposium on Computer Animation, the Outstanding Service Award as an Associate Editor of the Computers and Graphics journal by Elsevier, and several National Excellent Textbook Awards.
\end{IEEEbiography}
\vspace{-14pt}

\begin{IEEEbiography}[{\includegraphics[width=1in,height=1.25in,clip,keepaspectratio]{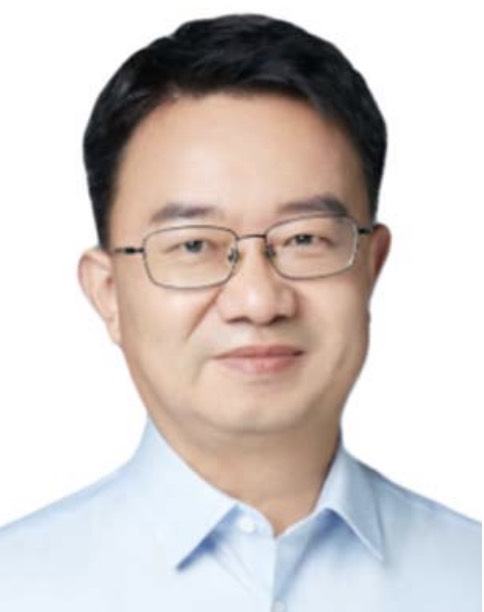}}]{Chuan Shi} received the BS degree from Jilin University in 2001, the MS degree from Wuhan University in 2004, and PhD degree from the ICT of Chinese Academic of Sciences in 2007. He is currently a professor with the Beijing University of Posts and Telecommunications. He has authored or coauthored 100 papers in refereed journals and conferences, such as SIGProc. 29th ACM SIGKDD Conf. Knowl. Discov. Data Mining, Proc. Int. Joint Conf. Artif. Intell., Proc.29th ACM Int. Conf. Inf. Knowl. Management, IEEE Transactions on Knowledge and Data Engineering, Proc. Int. Conf. World Wide WebJ, and ACM TIST. His research interests include data mining, machine learning, and evolutionary computing.
\end{IEEEbiography}
\vspace{-16pt}

\begin{IEEEbiography}[{\includegraphics[width=1in,height=1.25in,clip,keepaspectratio]{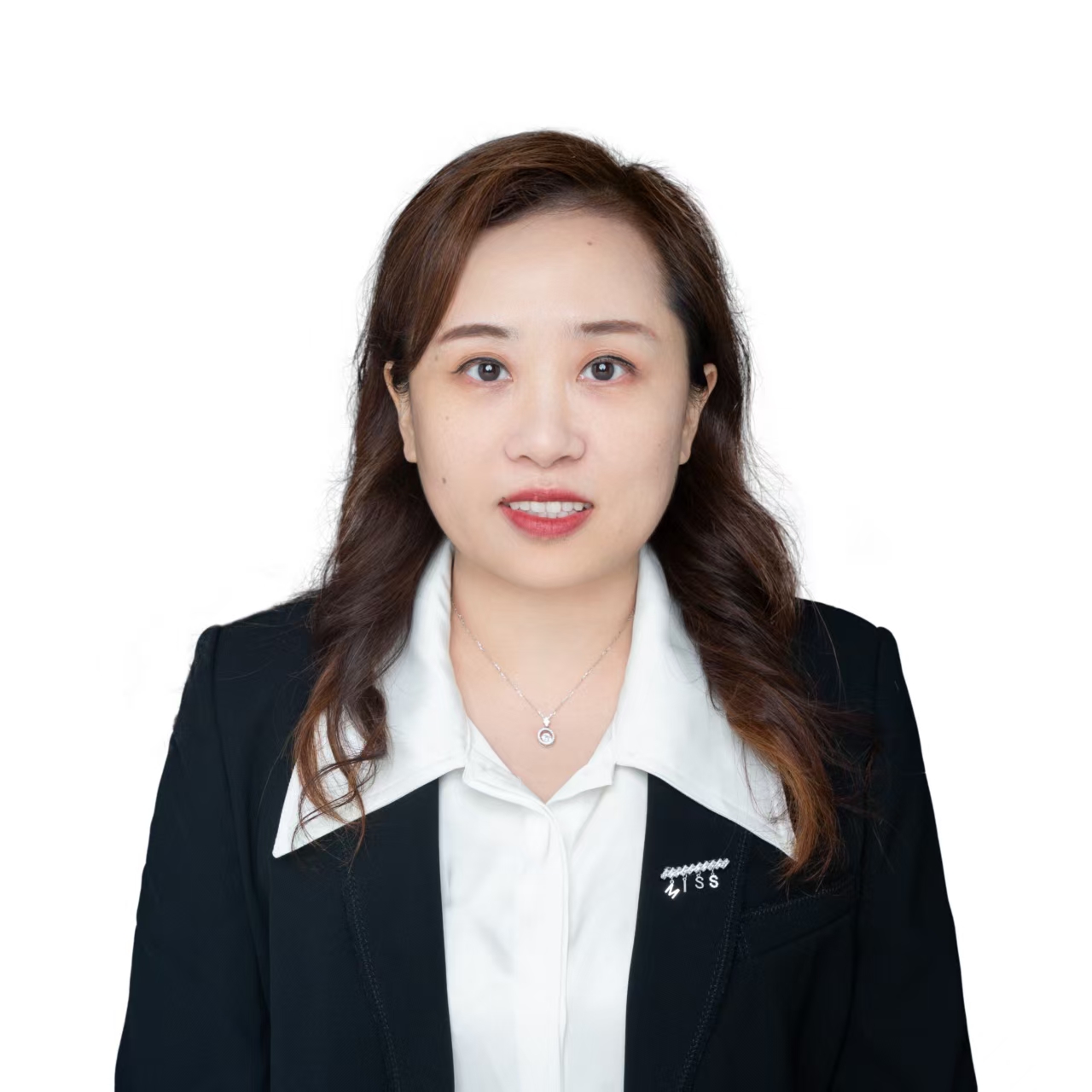}}]{Ling Tian}, Ph.D., professor
and doctoral supervisor. Her research interests mainly focus on knowledge graphs and knowledge reasoning.
\end{IEEEbiography}
\vspace{-16pt}

\begin{IEEEbiography}[{\includegraphics[width=1in,height=1.25in,clip,keepaspectratio]{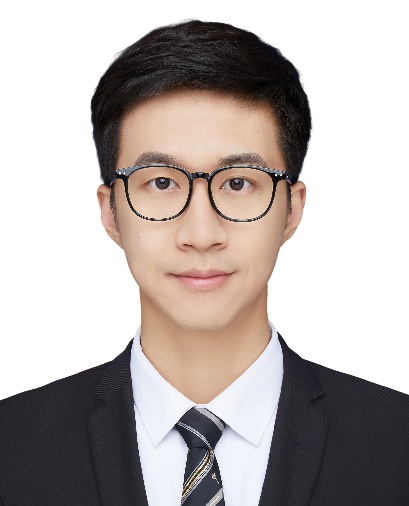}}]{Siqi Li} is a postdoctoral researcher with the School of Software, Tsinghua University. He received the B.S. degree from the Beihang University, Beijing, China, and the Ph.D. degree from Tsinghua University, Beijing, China. His research interests include computer vision and machine learning.
\end{IEEEbiography}
\vspace{-16pt}

\begin{IEEEbiography}[{\includegraphics[width=1in,height=1.25in,clip,keepaspectratio]{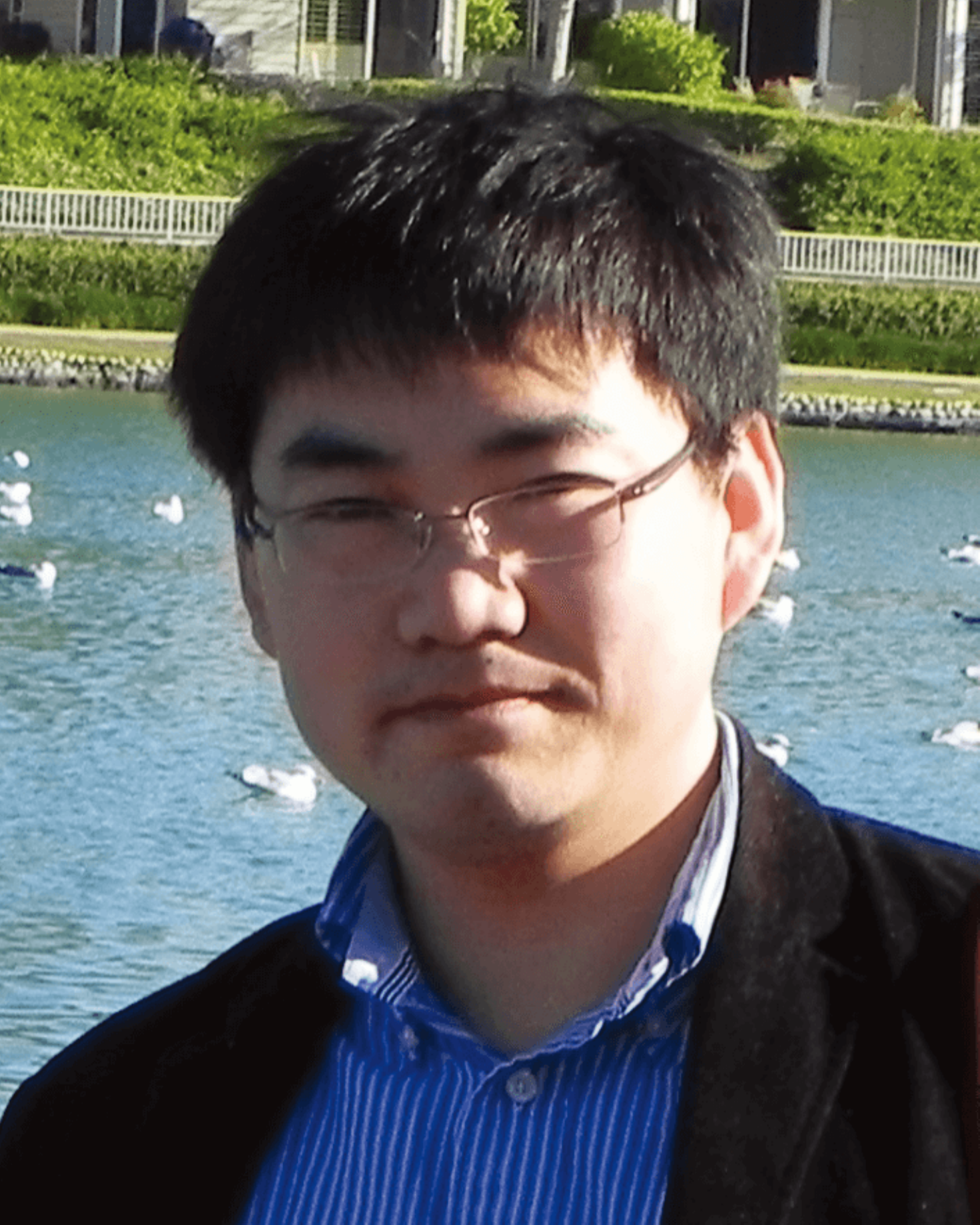}}]{Yue Gao} is an associate professor with the School of Software, Tsinghua University. He received the B.S. degree from the Harbin Institute of Technology, Harbin, China, and the M.E. and Ph.D. degrees from Tsinghua University, Beijing, China.
\end{IEEEbiography}

\vfill

\end{document}